\documentclass[pdflatex,sn-mathphys-num]{sn-jnl}

\usepackage{graphicx}%
\usepackage{multirow}%
\usepackage{amsmath,amssymb,amsfonts}%
\usepackage{amsthm}%
\usepackage{mathrsfs}%
\usepackage[title]{appendix}%
\usepackage{xcolor}%
\usepackage{textcomp}%
\usepackage{manyfoot}%
\usepackage{booktabs}%
\usepackage{algorithm}%
\usepackage{algorithmicx}%
\usepackage{algpseudocode}%
\usepackage{listings}%
\usepackage{lineno}
\usepackage{multirow}
\usepackage{subcaption}%
\usepackage{placeins}
\usepackage{booktabs}
\usepackage{float}
\usepackage{svg}
\usepackage{booktabs}
\usepackage{multirow}
\usepackage{graphicx}

\usepackage{svg} 
\usepackage{transparent-nometadata}

\theoremstyle{thmstyleone}%
\theoremstyle{thmstyletwo}%

\theoremstyle{thmstylethree}%

\begin{document}

\title[To Remove or Not to Remove Clouds]{To Remove or Not to Remove Clouds: A Comparative Analysis and Fusion of Raw SAR and Synthetic NDWI for Overcast Water Segmentation}

\author*[1]{\fnm{Saleh Sakib} \sur{Ahmed}}\email{salehsakibahmed@gmail.com}


\author*[2]{\fnm{Sara} \sur{Nowreen}}\email{snowreen@iwfm.buet.ac.bd}

\author*[1]{\fnm{M. Sohel} \sur{Rahman}}\email{msrahman@cse.buet.ac.bd}

\affil[1]{\orgdiv{Computer Science and Engineering}, \orgname{Bangladesh University of Engineering and Technology}, \orgaddress{\street{Palashi}, \city{Dhaka}, \country{Bangladesh}}}

\affil[2]{\orgdiv{Institute of Water and Flood Management  }, \orgname{Bangladesh University of Engineering and Technology}, \orgaddress{\street{Palashi}, \city{Dhaka}, \country{Bangladesh}}}

\abstract{
Persistent clouds blind optical satellites during floods. While Synthetic Aperture Radar (SAR) penetrates clouds, its raw data is noisy and lacks clear contrast. To mitigate this, recent studies utilize deep learning models to translate SAR into cloud-free synthetic optical imagery for downstream tasks like water body segmentation. However, because raw SAR is the original source for both of these operations, a critical methodological dilemma arises: during complete overcast should segmentation models process the raw SAR directly, or rely on a translated synthetic Normalized Difference Water Index (NDWI) proxy? This study resolves the debate by demonstrating that synthetic NDWI yields better results, as the translation process acts as a powerful filter against radar noise. This raises a natural second question: what if we utilize both? Building on our findings, we introduce a Combined Framework that integrates both raw SAR and synthetic NDWI into a unified model. By fusing the sharp physical boundaries of raw SAR with the high contrast of synthetic NDWI, this hybrid approach consistently outperforms all standalone methods.
}



\maketitle
\section{Introduction}
\label{sec:intro}

Optical satellite imagery is central to Earth observation, enabling the calculation of spectral indices that reveal critical surface features. The Normalized Difference Water Index (NDWI)~\cite{gao1996ndwi,mcfeeters1996use} is particularly essential for mapping water bodies and monitoring environmental extremes like floods~\cite{albertini2022detection,gu2007five} and droughts~\cite{chandrasekar2010land}. By exploiting water's high reflectance in the green band and strong absorption in the near-infrared (NIR) band, NDWI effectively isolates aquatic features. High-resolution optical missions, such as the European Space Agency's (ESA) Sentinel-2~\cite{copernicus_sentinel2}, drive these applications, supporting critical operations from disaster response~\cite{vlasova2023monitoring,tsyhanenko2025remote} to environmental conservation~\cite{ozelkan2020water}.

However, persistent cloud cover during emergencies like monsoons and cyclones renders optical data largely unavailable. To bypass this barrier, researchers utilize Synthetic Aperture Radar (SAR) satellites like Sentinel-1, as their microwave sensors easily penetrate clouds. Yet, raw SAR imagery suffers from inherent speckle noise and low contrast. Consequently, recent literature often focuses on translating SAR into synthetic, cloud-free optical imagery to facilitate diverse downstream tasks such as water body segmentation. 

Because raw SAR remains the foundational data source for both generating synthetic imagery and performing the downstream task, a critical question arises: should we introduce an intermediate step to synthesize optical images, or simply process the raw SAR directly? Focusing on water body segmentation, this research addresses this exact dilemma: \textit{Under completely overcast conditions, should models segment raw SAR directly, or rely on a translated synthetic NDWI proxy?} While studies~\cite{wieland2023s1s2} demonstrate that clean optical data outperforms raw SAR in cloud-free environments, no prior work has directly compared these two strategies when both must rely on the exact same underlying SAR source.

This study resolves this dilemma through a comprehensive comparative analysis of water body segmentation under total optical data unavailability. We demonstrate that synthetic NDWI yields significantly more accurate water extraction than the source raw radar data itself. Our analysis reveals that the SAR-to-optical translation process acts as a powerful structural filter, organizing chaotic radar noise into distinct features. Specifically, downstream segmentation accuracy strongly correlates with synthetic NDWI reconstruction quality ($R^2$). Additionally, we find that transfer learning-based models achieve superior SAR-to-optical mapping compared to custom-built architectures. Consequently, leveraging more powerful pretrained backbones improves the generation quality of synthetic NDWI, which directly yields higher segmentation accuracy.

Building on these insights, we ask a natural follow-up question: \textit{What if we use both data streams simultaneously?} Leveraging the principles of multi-view learning~\cite{li2018survey,yu2025review}—which posits that observing the same data from different perspectives provides richer context—we propose a novel \textbf{Combined Framework}. By simultaneously ingesting both modalities, this hybrid strategy fuses the sharp physical boundaries of raw SAR with the high contrast of synthetic NDWI, consistently outperforming both standalone approaches. 

Crucially, we intentionally restrict the translation target to a continuous, single-channel derived index (NDWI) rather than a full multi-spectral array. This lightweight intermediate representation minimizes computational overhead, enabling rigorous five-fold cross-validation and exhaustive ablation studies while establishing a robust, computationally efficient new standard for operational flood mapping under atmospheric blackout.
\section{Related Work}
\label{sec:related_work}

\subsection{Cloud Removal: Fusion vs. Translation}
\label{subsec:cloud_removal_literature}
Attempting to overcome atmospheric obstruction, existing literature relies on two main paradigms: multi-sensor fusion and SAR-to-optical translation. Fusion frameworks combine Sentinel-1 SAR and Sentinel-2 optical data to reconstruct missing optical pixels~\cite{cai2025fusing,meraner2020cloud,zhang2022removing,duan2024efficient,zhao2023seeing,ebel2020multisensor,anandakrishnan2024cermf,xu2022glf,jayakrishnan2023msdf}. However, fusion models exhibit two critical operational limitations during flood emergencies. First, \textit{temporal misalignment} creates a severe bottleneck, as Sentinel-1 and Sentinel-2 rarely capture identical coordinates simultaneously. Second, fusion models fail under \textit{total cloud cover}; without valid optical textures to fuse, the optical branch becomes redundant. Consequently, standalone SAR inputs remain the only operational choice under persistent overcast conditions.

To resolve this, generative deep learning models translate standalone radar into synthetic, cloud-free optical representations. Early Generative Adversarial Networks (GANs) focused on standard RGB synthesis~\cite{bermudez2018sar,fuentes2019sar,goodfellow2014generative}, whereas recent frameworks generate targeted spectral indices such as NDWI~\cite{ahmed2025light}. Modern architectures leverage advanced diffusion and flow-matching formulations, ranging from optical RGB reconstruction (e.g., Brownian Bridge Diffusion~\cite{kim2024conditional}) to multi-spectral array (including NDWI) synthesis (e.g., \textit{CloudBreaker}~\cite{ahmed2025cloudbreaker}). While these translation models aim to restore missing optical signals, they treat cross-modal generation purely as an end target rather than exploring its direct utility as a pre-filtering mechanism for downstream segmentation tasks.

\subsection{Water Segmentation Paradigms}
Remote sensing water body segmentation traditionally relies on three input paradigms: optical imagery~\cite{yuan2021deep, attya2025hybrid, shi2023improvement, jonnala2025aer}, standalone SAR data~\cite{pena2024deepaqua}, or multi-sensor fusion~\cite{declaro2024enhancing, wieland2023s1s2, bioresita2019fusion}. Optical approaches isolate water features using visible and infrared bands, while direct radar approaches rely strictly on microwave backscatter—the only signal consistently captured through heavy cloud cover. Traditional fusion streams attempt to join structural backscatter with optical spectral properties; however, they fundamentally depend on co-located optical inputs that are unavailable during severe overcast events.

Crucially, existing literature treats SAR-to-optical translation and radar-based segmentation as two isolated engineering pipelines. While multi-sensor fusion demonstrates clear performance gains in cloud-free conditions, no prior framework extends this multi-view concept to cloud-blocked scenarios by pairing raw SAR with synthetic spectral proxies. This study addresses this divide by empirically evaluating the synergistic relationship between cross-modal translation and downstream classification, establishing the foundational justification for our multi-view fusion framework under total optical data unavailability.

\begin{figure}
    \centering
    \includegraphics[width=0.95\linewidth]{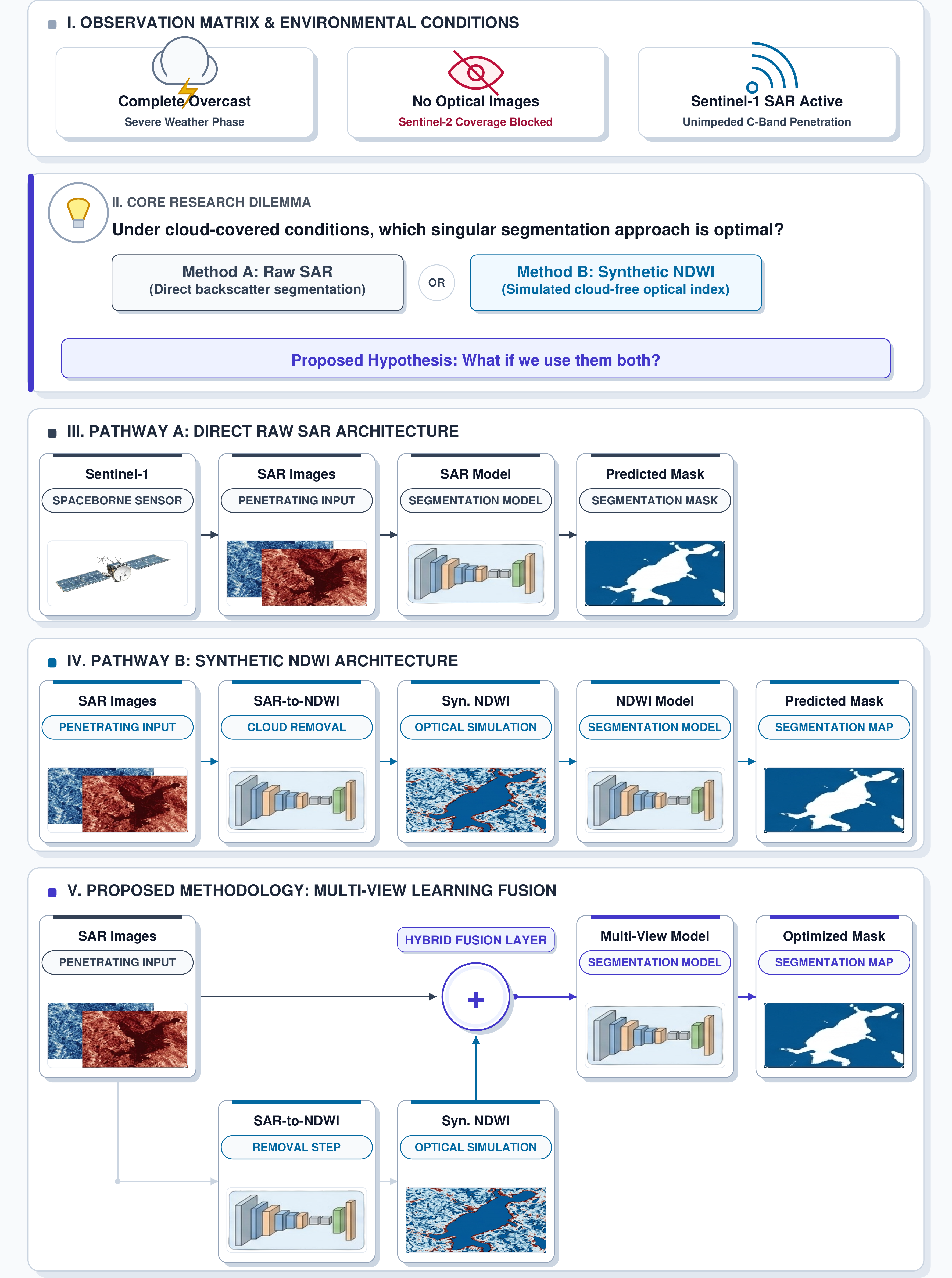}
    \caption{\textbf{Architectural paradigms for cloud-resilient water segmentation.} The figure depicts the main motivation of our research. It first presents the scenario and where this research is relevant, illustrating the challenge of surface monitoring under complete overcast conditions where Sentinel-2 optical imagery is completely blocked. Then, it presents the two conventional options: a direct raw SAR pipeline that bypasses cloud removal (Option 1) and a synthetic NDWI proxy pipeline that translates radar features into the optical domain (Option 2). Finally, our proposed method is presented at the bottom, showcasing a Multi-View Learning framework that simultaneously fuses raw SAR backscatter with the translated synthetic NDWI map to achieve robust, dual-stream feature extraction.}
    \label{fig:methodology_motivation}
\end{figure}

\section{Methodology}
\label{sec:methodology}
\begin{figure}
    \centering
    \includegraphics[width=\linewidth]{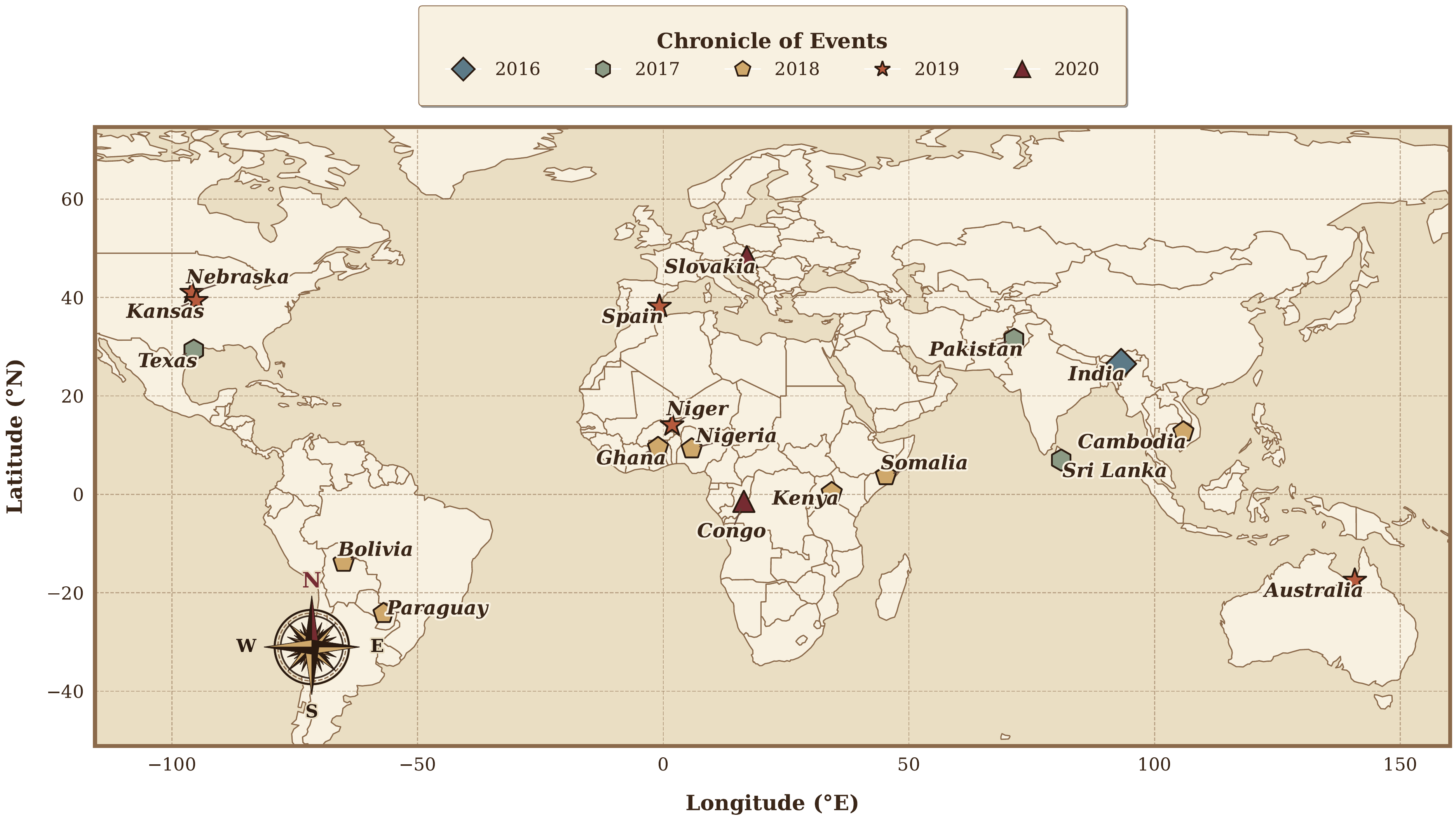}
    \caption{Global spatial and temporal distribution of the 18 flood events from the Microsoft Cloud2Street flood dataset (2016--2020). The map utilizes distinct colored geometric markers representing the location and specific year of each flood occurrence.}
    \label{fig:flood_event_map}
\end{figure}
\subsection{Datasets and Operational Data Roles}
\label{subsec:datasets_roles}

The \textit{Cloud to Street - Microsoft Flood and Clouds Dataset}~\cite{cloud2024} was utilized for this research. This data asset contains 900 co-registered pairs of Sentinel-1 (SAR) and Sentinel-2 (Optical) image chips ($512 \times 512$ pixels) spanning 18 global flood events. Sentinel-1 data includes the VV and VH polarization bands, while Sentinel-2 provides 13 spectral bands alongside native pixel-level water and cloud masks. Crucially, the target water masks are derived from Cloud to Street's global flood monitoring algorithms, providing standardized, algorithmically generated flood extent benchmarks across all event periods. To support the multi-stage deep learning framework, the individual data layers were designated for specific operational roles and formatted as follows:

\begin{description}
    \item[\textbf{Radar Input (Sentinel-1 SAR):}] The VV and VH polarization bands were stacked along the channel dimension to form a 3-dimensional tensor matching the chosen tile dimensions (either $128 \times 128$ or $256 \times 256$ pixels). To maintain numerical stability during training, all backscatter values were normalized linearly to a range between 0 and 1.
    \item[\textbf{Cloud Filter Mask (Sentinel-2 Cloud Mask):}] Native cloud and cloud shadow masks were utilized strictly during preprocessing to identify and eliminate obscured regions, ensuring that no corrupted atmospheric data influenced model evaluation.
    \item[\textbf{Synthetic NDWI Training Target (Sentinel-2 NDWI):}] The raw Green and NIR bands were used to compute continuous Normalized Difference Water Index (NDWI) maps according to Eq.~\ref{eq:NDWI}, serving as the regression ground truth for training the intermediate synthetic generator:
    \begin{equation}
    \label{eq:NDWI}
    \text{NDWI} = \frac{\text{Green} - \text{NIR}}{\text{Green} + \text{NIR}}
    \end{equation}
    While standard NDWI values inherently range from $-1$ to $1$, these values were linearly rescaled to a range between 0 and 1 during target preparation to match the bounded output domain of the model's final Sigmoid activation function.
    \item[\textbf{Segmentation Ground Truth (Native Water Mask):}] The algorithmically derived, pixel-level surface water extent masks provided natively within the dataset were utilized as the definitive binary targets to train and evaluate all final downstream segmentation streams. 
\end{description}

\subsection{Preprocessing}
\label{subsec:Preprocessing}
To optimize data handling and examine spatial scale trade-offs, the original $512 \times 512$ 900 image chips were systematically subdivided into smaller, non-overlapping patches (referred to as \textit{tiles}) under two distinct experimental configurations. In the high-data regime ($128 \times 128$), each chip was divided into 16 tiles to maximize training volume, yielding an initial raw pool of 14,400 samples. In the high-context regime ($256 \times 256$), each chip was divided into 4 tiles, reducing the total sample volume to 3,600 samples but preserving broader geographic and structural context.

A critical challenge in multimodal remote sensing is the temporal misalignment between Sentinel-1 and Sentinel-2 overpasses, which naturally leads to discrepancies in their respective ground truth masks due to dynamic surface water changes. To objectively resolve this and filter out tiles with significant temporal shifts, a data-driven thresholding approach was applied. Specifically, the pixel-wise disagreement percentage between the Sentinel-1 and Sentinel-2 water masks was calculated for each tile. By identifying the mathematical ``knee'' of the sorted disagreement distribution---calculated utilizing the maximum orthogonal distance from the curve to its secant line---a negligible difference threshold of 10.68\% for 128 was established which gave 12334 samples. 15.03\% for 256 giving 3229 samples.

Tiles exhibiting a disagreement above this threshold were discarded. This thresholding strategy achieved two major benefits: first, it functionally synchronized the two modalities by ensuring the recorded surface states were identical despite the acquisition time gap; second, it allowed the higher-quality Sentinel-2 water mask to serve as a unified, accurate ground truth for both sensors. This significantly minimized label noise for the Sentinel-1 to Sentinel-2 regression task.

Finally, a strict quality control constraint was applied to eliminate atmospheric interference. Any tile containing even a single cloud-shrouded pixel was discarded. Following this rigorous filtering pipeline, the final clean dataset resulted in 6,961 tiles for the $128 \times 128$ configuration and 1,552 tiles for the $256 \times 256$ configuration.

\subsection{Spatiotemporal Dataset Demographics}
\label{subsubsec:demographics}

Following the preprocessing and thresholding phases, the demographic composition of the retained dataset was analyzed across spatial and temporal dimensions to ensure adequate diversity for model generalization (summarized in Table~\ref{tab:dataset_demographics}). 

The $128 \times 128$ configuration retained tiles from all 18 original flood events, spanning 16 countries and 6 continents. Geographically, the dataset is predominantly driven by North America (34.49\%) and Africa (29.12\%). Temporally, the majority of the data captures events from the year 2019 (52.88\%) and is heavily concentrated in the transition months of September (23.65\%) and May (20.49\%). 

In contrast, the larger spatial footprint required for the $256 \times 256$ configuration resulted in a slightly narrower demographic spread. The strict cloud and temporal misalignment filtering caused two minor events (Bolivia and Somalia) to be entirely dropped from the high-context dataset, reducing the unique event count to 16. Despite these localized exclusions, the macro-level distribution remained highly consistent with the high-data regime, with North America (38.21\%) and the year 2019 (57.99\%) continuing to represent the majority of the training distribution.

\begin{table}[htpb]
    \centering
    \caption{Macro-level spatiotemporal distribution of the final dataset across the high-data ($128 \times 128$) and high-context ($256 \times 256$) regimes.}
    \label{tab:dataset_demographics}
   
    \begin{tabular}{@{}llrr|rr@{}}
        \toprule
        & & \multicolumn{2}{c}{\textbf{128 $\times$ 128 Regime}} & \multicolumn{2}{c}{\textbf{256 $\times$ 256 Regime}} \\
        \cmidrule(lr){3-4} \cmidrule(l){5-6}
        \textbf{Category} & \textbf{Attribute} & \textbf{Tile Count} & \textbf{Percentage (\%)} & \textbf{Tile Count} & \textbf{Percentage (\%)} \\
        \midrule
        \multirow{6}{*}{\textbf{Continent}} 
        & North America & 2,401 & 34.49 & 593 & 38.21 \\
        & Africa & 2,027 & 29.12 & 409 & 26.35 \\
        & Europe & 944 & 13.56 & 233 & 15.01 \\
        & Asia & 674 & 9.68 & 134 & 8.63 \\
        & South America & 654 & 9.40 & 141 & 9.09 \\
        & Oceania & 261 & 3.75 & 42 & 2.71 \\
        \midrule
        \multirow{5}{*}{\textbf{Year}} 
        & 2019 & 3,681 & 52.88 & 900 & 57.99 \\
        & 2018 & 1,220 & 17.53 & 245 & 15.79 \\
        & 2020 & 1,178 & 16.92 & 228 & 14.69 \\
        & 2017 & 691 & 9.93 & 146 & 9.41 \\
        & 2016 & 191 & 2.74 & 33 & 2.13 \\
        \midrule
        \multirow{5}{*}{\textbf{Top 5 Months}} 
        & September & 1,646 & 23.65 & 397 & 25.58 \\
        & May & 1,426 & 20.49 & 327 & 21.07 \\
        & August & 1,319 & 18.95 & 255 & 16.43 \\
        & October & 1,193 & 17.14 & 264 & 17.01 \\
        & April & 1,002 & 14.39 & 249 & 16.04 \\
        \midrule
        \textbf{Total Events} & & \textbf{18} & \textbf{100.00} & \textbf{16} & \textbf{100.00} \\
        \bottomrule
    \end{tabular}
\end{table}
\subsection{Cross-Validation and Geographic Stratification}
\label{subsec:cross_validation}

To rigorously evaluate model generalization across diverse global landscapes and prevent spatial data leakage, we implemented a 5-fold Stratified Group K-Fold cross-validation strategy. Instead of relying on random tile splits---which artificially inflate performance metrics due to the high spatial autocorrelation of adjacent image tiles---we grouped the dataset strictly by unique flood events. 

To ensure each training fold remained globally representative, events were first clustered into four spatial pseudo-continents based on their geographic coordinates. These spatial clusters acted as the strata during the K-Fold splitting process. This guaranteed two critical conditions across all five folds for both the $128 \times 128$ and $256 \times 256$ configurations:
\begin{enumerate}
    \item \textbf{Zero Geographic Leakage:} No tiles from a single flood event (e.g., a specific flood in Spain or Paraguay) were ever shared between the training and validation sets. The model was strictly evaluated on entirely unseen geographic regions.
    \item \textbf{Stratified Diversity:} Each training fold maintained a proportional representation of global regions, preventing the model from becoming biased toward the most data-heavy continents (such as North America or Africa).
\end{enumerate}

In the high-data ($128 \times 128$) configuration, the dataset encompassed diverse flooding events across 16 countries. The United States (34.49\%) and Niger (13.93\%) represented the largest data distributions, while smaller-scale localized events in Bolivia and Somalia represented the tail end (0.07\% each). 

When transitioning to the high-context ($256 \times 256$) configuration, the expanded spatial footprint and strict cloud-filtering constraints naturally eliminated the smallest localized events. Consequently, data from Bolivia and Somalia were entirely excluded from this regime. Despite this minor truncation, the global proportions remained largely stable, with the United States (38.21\%), Niger (15.21\%), and Paraguay (9.09\%) maintaining the majority of the training distribution, ensuring consistent baseline comparisons between the two scale regimes.
\subsection{Network Architecture: The Adapted SegFormer}
To use transfer learning while addressing the domain difference between SAR and optical imagery, the SegFormer framework was adopted \cite{xie2021segformer}. Specifically, Mix Transformer (MiT) encoders (ranging from MiT-B0 to MiT-B5) pretrained on ImageNet were used. Standard pretrained encoders expect a 3-channel RGB input, but Sentinel-1 data contains only 2 channels (VV/VH). To solve this, a \textbf{Learnable Input Adapter} was added. 

Let $X_{\text{SAR}} \in \mathbb{R}^{H \times W \times 2}$ represent the input SAR image. The adapter is a $1 \times 1$ convolutional layer $\mathcal{F}_{\text{adapt}}$ that projects the input into the required 3-channel feature space:
\begin{equation}
    X_{\text{RGB}} = \mathcal{F}_{\text{adapt}}(X_{\text{SAR}})
\end{equation}
This linear projection lets the pretrained encoder process the 2-channel SAR data without destroying the weights already learned from ImageNet in the very first layer.

For the regression task (SAR-to-NDWI generation), the network ends with a lightweight, All-MLP decoder followed by a bilinear upsampling operation to bring the output back to the original target image resolution. A Sigmoid activation function $\sigma(\cdot)$ is applied to the final output to keep the predicted pixel values between $0$ and $1$, matching the normalized target index.

\subsection{Optimization and Loss Functions}
\label{subsec:loss_functions}

The framework optimizes two distinct learning objectives: intermediate synthetic regression and downstream flood segmentation.

\subsubsection{Intermediate Regression Objective}
To train the regression network, a hybrid loss function handles noise in SAR data while preserving structural boundaries. The total regression loss $\mathcal{L}_{\text{reg}}$ is defined as:

\begin{equation}
    \mathcal{L}_{\text{reg}} = \alpha \cdot \mathcal{L}_{L1} + (1 - \alpha) \cdot \mathcal{L}_{\text{FocalMSE}}
\end{equation}

where the balancing parameter $\alpha$ is set to $0.6$. The standard $\mathcal{L}_{L1}$ loss provides robust pixel-level alignment, while the $\mathcal{L}_{\text{FocalMSE}}$ prioritizes complex coastlines and high-error regions by dynamically scaling training gradients based on the error magnitude:

\begin{equation}
    \mathcal{L}_{\text{FocalMSE}} = \frac{1}{N} \sum_{i=1}^{N} (|y_i - \hat{y}_i| + \epsilon)^\gamma (y_i - \hat{y}_i)^2
\end{equation}

where $y_i$ is the ground truth feature, $\hat{y}_i$ is the prediction, $\gamma=0.8$ controls focusing strength to prevent background pixel dominance, and $\epsilon=10^{-6}$ ensures numerical stability.

\subsubsection{Downstream Segmentation Objective}
To address severe class imbalance in flood extents, the downstream segmentation network is optimized using a combination of Positional-Weighted Binary Cross-Entropy with Logits ($\mathcal{L}_{\text{BCE}}$) and Soft Dice Loss ($\mathcal{L}_{\text{Dice}}$):

\begin{equation}
    \mathcal{L}_{\text{seg}} = w_{\text{BCE}} \cdot \mathcal{L}_{\text{BCE}} + w_{\text{Dice}} \cdot \mathcal{L}_{\text{Dice}}
\end{equation}

where $w_{\text{BCE}} = 1.0$ and $w_{\text{Dice}} = 1.0$. Given network logits $x$ and binary targets $t$, the constituent terms are formalized as:

\begin{equation}
    \mathcal{L}_{\text{BCE}} = -\frac{1}{N} \sum_{i=1}^{N} \left[ w_{\text{pos}} \cdot t_i \cdot \log \sigma(x_i) + (1 - t_i) \cdot \log (1 - \sigma(x_i)) \right]
\end{equation}

\begin{equation}
    \mathcal{L}_{\text{Dice}} = 1 - \frac{2 \sum_{i=1}^{N} \sigma(x_i) t_i + \text{smooth}}{\sum_{i=1}^{N} \sigma(x_i) + \sum_{i=1}^{N} t_i + \text{smooth}}
\end{equation}

where $\sigma(\cdot)$ represents the Sigmoid activation function, $w_{\text{pos}} = 2.0$ scales positive water pixel gradients, and $\text{smooth} = 10^{-6}$ prevents division-by-zero during tensor flattening.

\subsection{Experimental Strategy: Three-Stream Evaluation}
To evaluate the downstream usefulness of the synthetic optical features, a comparative study was developed using three distinct segmentation streams trained via 5-Fold Cross-Validation:

\begin{enumerate}
    \item \textbf{Baseline (S1 Only):} A standard segmentation model trained directly on the original 2-channel SAR input ($X_{\text{SAR}} \to Y_{\text{mask}}$). This serves as the baseline reference for performance.
    \item \textbf{Synthetic Branch (Synthetic S2 Only):} A segmentation model trained only on the generated NDWI images ($\hat{X}_{\text{S2}} \to Y_{\text{mask}}$). This isolates the quality of the regression generator; a high-quality synthetic image should allow this model to achieve results close to or better than the raw radar baseline.
    \item \textbf{Combined Framework (Proposed):} A fusion model where the original SAR data is combined with the synthetic optical features along the channel dimension ($[X_{\text{SAR}}; \hat{X}_{\text{S2}}]$). This strategy allows the network to use the underlying physical structures of SAR while using the translated spectral clues to resolve radar errors.
\end{enumerate}

\paragraph{Justification for Feature Fusion}
\label{para:justification}
The reason for training on both the original data ($S_1$) and its synthetic derivative ($S_2$) is based on the principle of multi-view learning~\cite{li2018survey,yu2025review}, where different versions of the same signal help the model extract more reliable features. While $S_1$ provides real-world grounding by keeping the authentic noise and complexity of radar data, $S_2$ acts as a clean or structured version of that same information. By processing both the raw, noisy data ($S_1$) and the cleaner, structured version ($S_2$), the model is forced to learn stable representations that ignore minor surface errors. This dual exposure acts as a regularizer during training \cite{diligenti2017semantic}: it prevents the model from overfitting to the quirks of the original radar source while using the clarity of the synthetic image to sharpen its predictions, leading to a model that is both grounded in reality and stable on new data.

\subsection{Implementation Details}
The framework was implemented using PyTorch and the Hugging Face Transformers library. Training was performed using the AdamW optimizer with a weight decay of $1 \times 10^{-2}$. The \textit{OneCycleLR} scheduler was utilized to warm up the learning rate to a maximum of $3 \times 10^{-4}$ before lowering it, ensuring stable convergence. To optimize training speed and memory usage, Automatic Mixed Precision (AMP) was employed. For the regression stage, the $R^2$ score (Coefficient of Determination) was monitored on the validation set, and early stopping with a patience of 40 epochs was used to prevent overfitting.
\begin{figure}[H]
\begin{center}
   
    \label{fig:sar_vs_s2_comparison}
\end{center}

\end{figure}
\section{Results}
\label{sec:results}

This section presents the empirical evaluation of deep learning paradigms for surface water segmentation under persistent cloud cover where only Synthetic Aperture Radar (SAR) data is accessible. As detailed in Section~\ref{subsec:datasets_roles}, the experimental data comprises paired Sentinel-1 SAR backscatter profiles and Sentinel-2 optical imagery, supplemented by automated cloud masks for overcast filtering and verified surface water maps serving as the segmentation ground truth.

To track performance comprehensively from intermediate feature translation to final mapping, the evaluation pipeline is structured into four sequential phases:
\begin{enumerate}
    
    \item \textbf{SAR-to-NDWI Translation:} An intermediate SAR-to-NDWI translation analysis is conducted between models trained from scratch and cross-domain pretrained architectures for synthetic NDWI generation.
    \item \textbf{Modality Gap Baseline:} A baseline is established to reaffirm the performance gap between clean, real Sentinel-2 optical indices and raw Sentinel-1 SAR data.
    \item \textbf{Downstream Segmentation:} Downstream segmentation accuracy is evaluated across standalone SAR, purely synthetic optical, and fused hybrid configurations.
    \item \textbf{Correlation and Ablation Testing:} The direct correlation between intermediate regression fidelity and downstream segmentation performance is analyzed, complemented by ablation testing across the two spatial tile sizes defined in Section~\ref{subsec:Preprocessing}.
\end{enumerate}

\subsection{Phase 1: Evaluation of Intermediate Synthetic NDWI Generation}
\label{subsec:phase1_regression}

\subsubsection{Comparative Analysis of Pretrained and From-Scratch Architectures}
\label{subsubsec:comparative_scratch_pretrained}
As an intermediate objective of this architecture, the performance of various models for generating synthetic NDWI was evaluated to optimize the translation stage. A comparative analysis was conducted between established architectures (e.g., U-Net, ViT, and Swin Transformer) initialized with random weights and the SegFormer family utilizing ImageNet-pretrained encoders.

As summarized in Table~\ref{table:performance}, a decisive performance gap favors the transfer learning configurations across most key metrics. The best-performing model trained from scratch in terms of explained variance, a deep U-Net (depth=5, groups=4), achieved an $R^2$ score of $0.7609$ but required $124.39$M parameters. In contrast, the lightest pretrained configuration, SegFormer (mit-b0), achieved a higher $R^2$ score of $0.7745$ while using only $3.70$M parameters—representing a $97\%$ reduction in parameter burden compared to the heaviest scratch U-Net variant. While the scratch U-Net (depth=4, groups=8) retained a marginal advantage in structural mapping (peak SSIM of $0.6933$), the pretrained models dominated in overall error reduction. Within the transfer learning group, the highest absolute validation performance was achieved by the SegFormer (mit-b5) backbone, reaching an optimal $R^2$ score of $0.8269$, a minimal MSE of $0.0076$, and a wMAPE of $22.21\%$. 

Consequently, pretrained Transformer encoders were selected as the definitive backbone for high-fidelity synthetic NDWI generation within this framework. A comprehensive discussion interpreting these findings—specifically detailing the mechanisms of cross-domain adaptation, feature generalization, and parameter efficiency—is provided in Section~\ref{sec:supp_pretraining} of the Supplementary Material.

\begin{table*}[!ht]
\centering
\caption{Comparison of synthetic NDWI generation performance. The evaluation isolates architectures trained from scratch against variants utilizing cross-domain pretraining. Bold text denotes the best overall metric, and underlined text denotes the second best.}
\label{table:performance}
\begin{tabular}{p{3.25cm}rrrrrrp{2cm}}
\toprule
Model & Params (M) & MSE & MAE & R$^2$ & SSIM & wMAPE (\%) & Remarks \\
\midrule
\multicolumn{8}{c}{\textbf{--- Models Trained from Scratch ---}} \\
\midrule
U-Net \\(depth=3, groups=4) & \underline{7.70} & 0.0107 & 0.0710 & 0.7560 & 0.6899 & 27.77 & \multirow{3}{*}{Ronneberger et al. \cite{ronneberger2015u}} \\
U-Net \\(depth=4, groups=8) & 31.04 & 0.0107 & 0.0700 & 0.7550 & \textbf{0.6933} & 27.38 &  \\
U-Net \\(depth=5, groups=4) & 124.39 & 0.0104 & 0.0707 & 0.7609 & \underline{0.6913} & 27.65 &  \\
\addlinespace
ViT (depth=8) & 12.38 & 0.0149 & 0.0832 & 0.6578 & 0.6523 & 32.55 & \multirow{2}{*}{Dosovitskiy et al. \cite{dosovitskiy2020image}} \\
ViT (depth=16) & 24.75 & 0.0479 & 0.1424 & -0.0965 & 0.5678 & 55.73 &  \\
\addlinespace
Swin Transformer (Tiny) & 16.14 & 0.0109 & 0.0709 & 0.7504 & 0.6885 & 27.76 & Liu et al. \cite{liu2021swin} \\
\midrule
\multicolumn{8}{c}{\textbf{--- Models using Pretraining (Transfer Learning) ---}} \\
\midrule
SegFormer (mit-b0) & \textbf{3.70} & 0.0098 & 0.0666 & 0.7745 & 0.6605 & 26.08 & \multirow{3}{*}{Xie et al. \cite{xie2021segformer}} \\
SegFormer (mit-b3) & 44.60 & \underline{0.0080} & \underline{0.0581} & \underline{0.8163} & 0.6753 & \underline{22.72} &  \\
SegFormer (mit-b5) & 82.00 & \textbf{0.0076} & \textbf{0.0568} & \textbf{0.8269} & 0.6764 & \textbf{22.21} &  \\
\bottomrule
\end{tabular}
\end{table*}

\subsubsection{Capacity Scaling of Pretrained Synthetic NDWI Generators}
Building upon the validation in Section~\ref{subsubsec:comparative_scratch_pretrained}, cross-validation capacity scaling was extended across both spatial data regimes using Mix Transformer (\textbf{MiT}) encoders. As detailed in Table~\ref{table:regression_scaling}, a consistent positive scaling trend is observed within the high-data regime ($128 \times 128$). The largest configuration, \textbf{MiT-B5}, reaches an optimal $R^2$ score of $0.8079 \pm 0.01$ and a minimal Mean Squared Error (MSE) of $0.0084 \pm 0.00$. 

Conversely, inside the high-context footprint ($256 \times 256$), scaling the capacity from MiT-B0 to \textbf{MiT-B3} unlocks substantial structural mapping performance, driving the $R^2$ score from $0.7390$ to $0.7531$ and producing an optimal Structural Similarity Index (SSIM) of $0.6818 \pm 0.00$. Notably, scaling further to \textbf{MiT-B5} at this resolution resulted in a performance plateau and slight degradation (with the $R^2$ score dropping to $0.7481 \pm 0.03$ and wMAPE increasing to $22.83\%$). This indicates that for the lower-data, high-context $256 \times 256$ regime, MiT-B3 provides the optimal representational capacity before overfitting or optimization bottlenecks occur. This might be due to the lower number of data points present in 256 tile size rather than 128. As more tiles were filtered in 256 resulting in lower number of training points.
\begin{table*}[!ht]
\centering
\caption{Cross-validation regression scaling performance metrics categorized across spatial data resolutions.}
\label{table:regression_scaling}
\setlength{\tabcolsep}{10pt}
\begin{tabular}{l|ccc|cc}
    \toprule
    \multirow{2}{*}{\textbf{Metric}} & \multicolumn{3}{c|}{\textbf{128 $\times$ 128 Tiles (High Data)}} & \multicolumn{2}{c}{\textbf{256 $\times$ 256 Tiles (High Context)}} \\
    \cmidrule{2-6}
     & \textbf{mit-b0} & \textbf{mit-b3} & \textbf{mit-b5} & \textbf{mit-b0} & \textbf{mit-b3} \\
    \midrule
    $R^2$ Score $\uparrow$ & 0.7835 $\pm$ 0.01 & 0.7956 $\pm$ 0.02 & \textbf{0.8079 $\pm$ 0.01} & 0.7390 $\pm$ 0.04 & \textbf{0.7531 $\pm$ 0.05} \\
    MSE $\downarrow$ & 0.0094 $\pm$ 0.00 & 0.0089 $\pm$ 0.00 & \textbf{0.0084 $\pm$ 0.00} & 0.0090 $\pm$ 0.00 & \textbf{0.0086 $\pm$ 0.00} \\
    MAE $\downarrow$ & 0.0642 $\pm$ 0.00 & 0.0608 $\pm$ 0.00 & \textbf{0.0593 $\pm$ 0.00} & 0.0653 $\pm$ 0.01 & \textbf{0.0626 $\pm$ 0.01} \\
    SSIM $\uparrow$ & 0.6614 $\pm$ 0.01 & 0.6689 $\pm$ 0.01 & \textbf{0.6702 $\pm$ 0.01} & 0.6739 $\pm$ 0.01 & \textbf{0.6818 $\pm$ 0.00} \\
    wMAPE (\%) $\downarrow$ & 25.17 $\pm$ 0.53 & 23.82 $\pm$ 0.69 & \textbf{23.21 $\pm$ 0.65} & 23.47 $\pm$ 1.08 & \textbf{22.54 $\pm$ 1.64} \\
    LPIPS $\downarrow$ & 0.5733 $\pm$ 0.01 & 0.5688 $\pm$ 0.01 & \textbf{0.5569 $\pm$ 0.01} & 0.6430 $\pm$ 0.02 & \textbf{0.6133 $\pm$ 0.02} \\
    FID $\downarrow$ & \textbf{4.2094 $\pm$ 0.17} & 4.2538 $\pm$ 0.18 & 4.2370 $\pm$ 0.13 & \textbf{8.5230 $\pm$ 0.73} & 8.5232 $\pm$ 0.80 \\
    \bottomrule
\end{tabular}
\end{table*}

\begin{figure}[htbp]
    \centering
    \includegraphics[width=\linewidth]{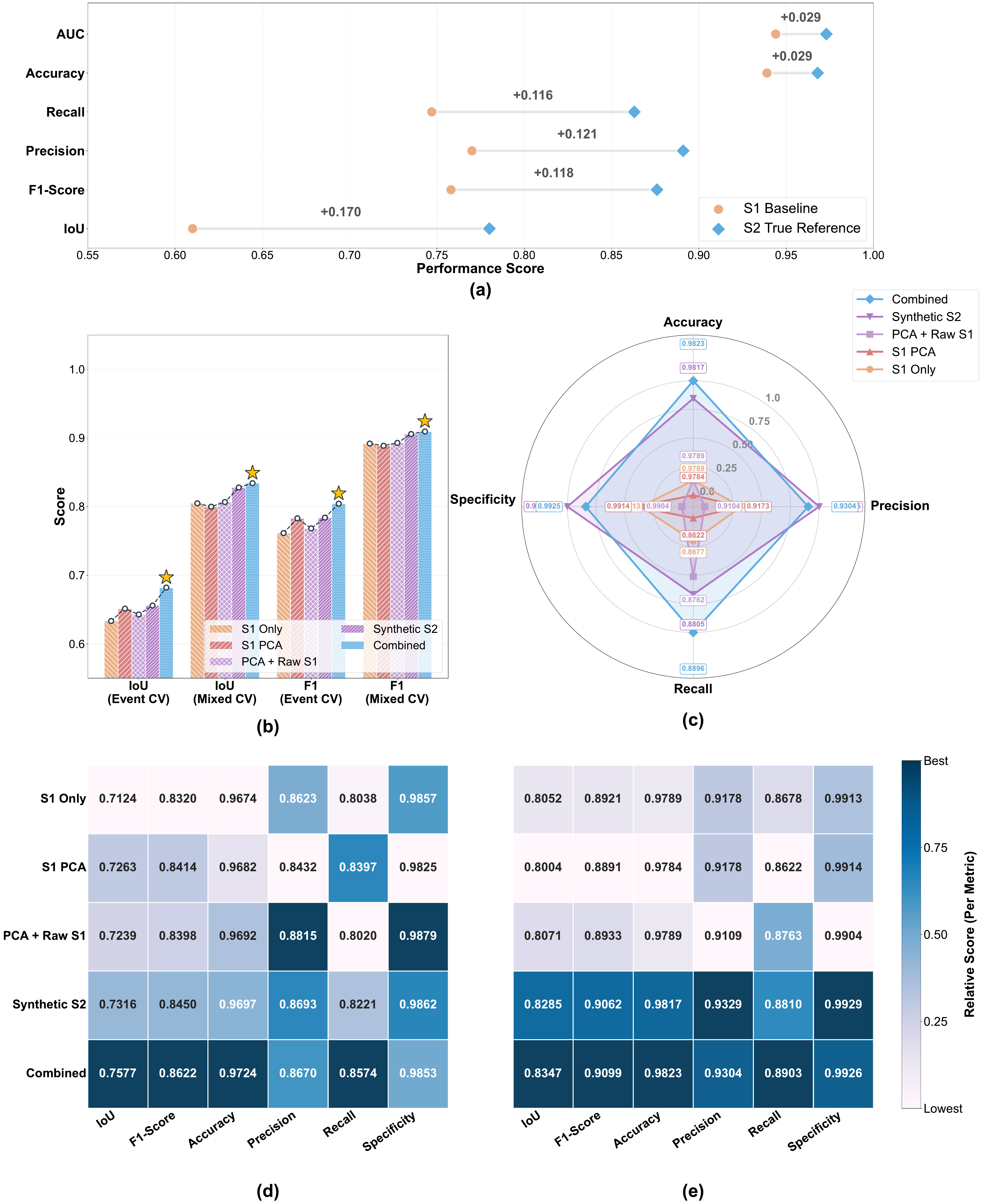}
    \caption{\textbf{Comprehensive performance evaluation of the proposed fusion and translation strategies.} \textbf{(a)} Modality gap analysis illustrating the baseline performance disparity between S1 (Baseline) and S2 (True Reference) across key segmentation metrics. \textbf{(b)} Evaluation of IoU and F1-Scores showing averaged Cross-Validation (CV) results. Two data splitting paradigms are compared: Event-stratified (where data is split by distinct discrete events to evaluate generalization) and Mixed (where data is uniformly shuffled across all events prior to splitting). The dashed line tracks the performance trend, with golden stars explicitly denoting the highest-scoring configuration for each metric. \textbf{(c)} Normalized radar plot providing a multidimensional comparison of Accuracy, Precision, Recall, and Specificity to highlight trade-offs between the strategies. \textbf{(d, e)} Normalized heatmaps detailing the relative Global performance (metrics computed over the entire aggregated test set rather than averaged per-fold) of each strategy under Event (d) and Mixed (e) data splitting paradigms; darker shades of blue indicate superior relative performance for a given metric.}
    \label{fig:performance_evaluation}
\end{figure}
\begin{table*}[!ht]
\centering
\caption{Global and cross-validation segmentation performance alongside paired statistical significance. The segmentation network utilizes SegFormer (MiT-B0), and the synthetic representations are generated using the optimal MiT-B5 regression model. Best results among the proposed strategies are \textbf{bolded}, and second-best are \underline{underlined} (excluding the Original S2 Oracle). Panels C and D report Wilcoxon signed-rank tests with a defined significance threshold of $p < 0.05$.}
\label{table:main_results}

\begin{tabular}{l c c c c}
\toprule
\multirow{2}{*}{\textbf{Input Modality}} & \multicolumn{2}{c}{\textbf{IoU}} & \multicolumn{2}{c}{\textbf{F1-Score}} \\
\cmidrule(lr){2-3} \cmidrule(lr){4-5}
& \textbf{Global} & \textbf{CV (Mean $\pm$ SD)} & \textbf{Global} & \textbf{CV (Mean $\pm$ SD)} \\
\midrule
\multicolumn{5}{c}{\textbf{Panel A: Event Stratified Splitting Performance}} \\
\midrule
S1 Only & 0.7124 & 0.6334 $\pm$ 0.1710 & 0.8320 & 0.7615 $\pm$ 0.1352 \\
S1 PCA & 0.7263 & 0.6514 $\pm$ 0.1192 & 0.8414 & 0.7830 $\pm$ 0.0820 \\
S1 PCA + Raw S1 & 0.7239 & 0.6429 $\pm$ 0.1733 & 0.8398 & 0.7683 $\pm$ 0.1372 \\
Synthetic S2 (Reg) & \underline{0.7316} & \underline{0.6559 $\pm$ 0.1395} & \underline{0.8450} & \underline{0.7840 $\pm$ 0.0975} \\
Combo (Synth S2 + S1) & \textbf{0.7577} & \textbf{0.6821 $\pm$ 0.1282} & \textbf{0.8622} & \textbf{0.8043 $\pm$ 0.0886} \\
\midrule
\textit{Original S2 (Oracle)} & \textit{0.8406} & \textit{0.8209 $\pm$ 0.0740} & \textit{0.9134} & \textit{0.8998 $\pm$ 0.0463} \\
\midrule
\multicolumn{5}{c}{\textbf{Panel B: Mixed Splitting Performance}} \\
\midrule
S1 Only & 0.8052 & 0.8049 $\pm$ 0.0100 & 0.8921 & 0.8919 $\pm$ 0.0061 \\
S1 PCA & 0.8004 & 0.7999 $\pm$ 0.0189 & 0.8891 & 0.8887 $\pm$ 0.0116 \\
S1 PCA + Raw S1 & 0.8071 & 0.8067 $\pm$ 0.0140 & 0.8933 & 0.8929 $\pm$ 0.0086 \\
Synthetic S2 (Reg) & \underline{0.8285} & \underline{0.8279 $\pm$ 0.0154} & \underline{0.9062} & \underline{0.9058 $\pm$ 0.0093} \\
Combo (Synth S2 + S1) & \textbf{0.8347} & \textbf{0.8342 $\pm$ 0.0164} & \textbf{0.9099} & \textbf{0.9095 $\pm$ 0.0099} \\
\midrule
\textit{Original S2 (Oracle)} & \textit{0.8689} & \textit{0.8683 $\pm$ 0.0103} & \textit{0.9298} & \textit{0.9295 $\pm$ 0.0059} \\
\bottomrule
\end{tabular}

\vspace{1.5em} 

\begin{tabular}{l c c c}
\toprule
\textbf{Hypothesis Comparison (Test $>$ Baseline)} & \textbf{Statistic} & \textbf{$p$-value} & \textbf{Significance} \\
\midrule
\multicolumn{4}{c}{\textbf{Panel C: Event Stratified Splitting Statistical Significance}} \\
\midrule
S1 PCA $>$ S1 Only & 1,250,140.5 & $2.02 \times 10^{-11}$ & \textbf{Significant} \\
S1 PCA + Raw S1 $>$ S1 Only & 1,348,625.5 & $1.79 \times 10^{-31}$ & \textbf{Significant} \\
Synthetic S2 (Reg) $>$ S1 PCA & 1,213,819.0 & $1.62 \times 10^{-3}$ & \textbf{Significant} \\
Combo (Synth S2 + S1) $>$ S1 PCA & 1,768,736.5 & $8.99 \times 10^{-57}$ & \textbf{Significant} \\
Combo (Synth S2 + S1) $>$ S1 PCA + Raw S1 & 1,724,481.0 & $5.59 \times 10^{-44}$ & \textbf{Significant} \\
\midrule
\multicolumn{4}{c}{\textbf{Panel D: Mixed Splitting Statistical Significance}} \\
\midrule
S1 PCA $>$ S1 Only & 1,187,596.0 & $0.9999$ & Not Sig. \\
S1 PCA + Raw S1 $>$ S1 Only & 1,356,024.0 & $0.0648$ & Not Sig. \\
Synthetic S2 (Reg) $>$ S1 PCA & 1,788,832.0 & $1.51 \times 10^{-29}$ & \textbf{Significant} \\
Combo (Synth S2 + S1) $>$ S1 PCA & 2,126,303.5 & $1.13 \times 10^{-87}$ & \textbf{Significant} \\
Combo (Synth S2 + S1) $>$ S1 PCA + Raw S1 & 2,006,882.5 & $1.86 \times 10^{-60}$ & \textbf{Significant} \\
\bottomrule
\end{tabular}

\end{table*}

\subsection{Baseline Performance Analysis: Modality Gap Reaffirmation (S1 vs.\ S2)}
\label{subsec:modality_gap}

Extensive prior literature has established the inherent superiority of Sentinel-2 (S2) optical data over Sentinel-1 (S1) Synthetic Aperture Radar (SAR) for water body segmentation. Although true S2 optical information is practically unavailable under completely overcast conditions, it is necessary to first reaffirm this modality gap on our specific dataset to establish a clear performance ceiling. To quantify the baseline difficulty of relying solely on SAR data, an empirical upper bound was established using cloud-free S2 references and systematically compared against a standalone S1 baseline.

When shifting from optical to radar data, a significant performance degradation across all metrics is revealed by the quantitative results visualized in Figure~\ref{fig:performance_evaluation}a. The performance ceiling is defined by the S2 optical reference (True Reference), which achieves a Global Intersection over Union (IoU) of \textbf{0.8406} and an F1-score of \textbf{0.9134} under the Event Stratified evaluation. In contrast, a sharp drop to a Global IoU of \textbf{0.7124} and an F1-score of \textbf{0.8320} is observed for the standalone S1 baseline. This multi-metric deficit is visualized directly in the modality gap analysis (Fig.~\ref{fig:performance_evaluation}a), demonstrating a severe collapse across all tracked dimensions, including Precision, Recall, Accuracy, and Specificity. 

These distinct drops in performance, particularly in IoU and Recall, are fundamentally driven by the physical limitations of radar imaging, such as speckle noise and terrain-induced geometric distortions (e.g., radar shadow and layover) that obscure land-water boundaries. Ultimately, this local verification of the modality gap precisely quantifies the performance deficit that the proposed generative framework aims to bridge by translating raw, noise-degraded SAR streams into clean, synthetic optical feature spaces.

\subsection{Phase 2: Downstream Main Segmentation Results}
\label{subsec:phase2_segmentation}

The core objective of this phase is to evaluate the downstream segmentation performance across different processing configurations. Two primary architectural stages are involved: the regression stage (synthetic NDWI continuous feature translation) and the segmentation stage (binary water body delineation). While the baseline S1-only configuration maps surface water directly from raw radar backscatter data, both the Synthetic S2-only baseline and the proposed Combined framework utilize both stages sequentially. 

To convey these structural combinations clearly, configurations are denoted as \textbf{(Reg-[X] / Seg-[Y])}, where \textbf{Reg-[X]} indicates the capacity of the intermediate regression network, and \textbf{Seg-[Y]} represents the downstream encoder capacity used for binary mask generation.

The primary configuration selected for this evaluation is \textbf{(Reg-B5 / Seg-B0)}, which pairs a heavy MiT-B5 regression feature extractor with a lightweight, parameter-efficient MiT-B0 segmentation head. This hybrid model is systematically benchmarked against two standalone configurations: (i) \textbf{Synthetic S2 Only}, which segments purely from the synthetic index space and discards the physical radar signals, and (ii) \textbf{S1 Only}, which directly processes raw backscatter data without an intermediate translation or cloud-removal step. Additional PCA-based fusion baselines (S1 PCA and S1 PCA + Raw S1) are also included for comprehensive benchmarking to compare the proposed generative translation against traditional statistical dimensionality reduction.

\subsubsection{Mixed and Event-Stratified Evaluation}
As demonstrated by the quantitative metrics in Table~\ref{table:main_results} and Figure~\ref{fig:performance_evaluation}b, the multi-view \textbf{Combined (Synth S2 + S1)} framework consistently achieves superior performance. Under the Mixed splitting scheme, the PCA-based baselines offer minimal to no benefit over the raw radar data; \textbf{S1 PCA} (IoU: $0.7999$) and \textbf{S1 PCA + Raw S1} (IoU: $0.8067$) perform similarly to the \textbf{S1 Only} baseline (IoU: $0.8049$), with Wilcoxon signed-rank tests confirming these minor differences are not statistically significant ($p > 0.05$, Table~\ref{table:main_results} Panel D). In contrast, the Combined approach yields a cross-validation mean IoU of \textbf{0.8342 $\pm$ 0.0164} and an F1-Score of \textbf{0.9095 $\pm$ 0.0099}, significantly outperforming the S1 baselines and the translation-only Synthetic S2 alternative (IoU: $0.8279$, F1: $0.9058$). 

This consistent margin of improvement is even more pronounced under the rigorous Event-stratified cross-validation paradigm (designed to test generalization across distinct geographic events). Here, traditional feature fusion provides a measurable benefit, with both \textbf{S1 PCA} (IoU: $0.6514$) and \textbf{S1 PCA + Raw S1} (IoU: $0.6429$) achieving statistically significant improvements ($p < 0.001$) over the \textbf{S1 Only} model ($0.6334 \pm 0.1710$). However, the generative translation of the \textbf{Combined} framework reaches an IoU of \textbf{0.6821 $\pm$ 0.1282}, continuing to substantially outperform all S1 and PCA-based baselines. As reported in Panels C and D of Table~\ref{table:main_results}, statistical tests confirm that the performance gains of the Combined framework over both the Synthetic S2 and PCA fusion baselines are highly significant ($p < 0.05$) across all evaluated configurations. 

Furthermore, the normalized radar plot (Fig.~\ref{fig:performance_evaluation}c) and relative performance heatmaps (Fig.~\ref{fig:performance_evaluation}d, e) highlight that this superiority is multidimensional. The Combined strategy effectively balances Precision and Recall (Table~\ref{table:supp_metrics_4panel}), confirming that simultaneous optimization over physical radar boundaries and translated optical feature spaces yields a more stable and accurate representation than tracking either individual modality alone.

\begin{figure}[!htbp]
    \centering
    \includegraphics[width=\linewidth]{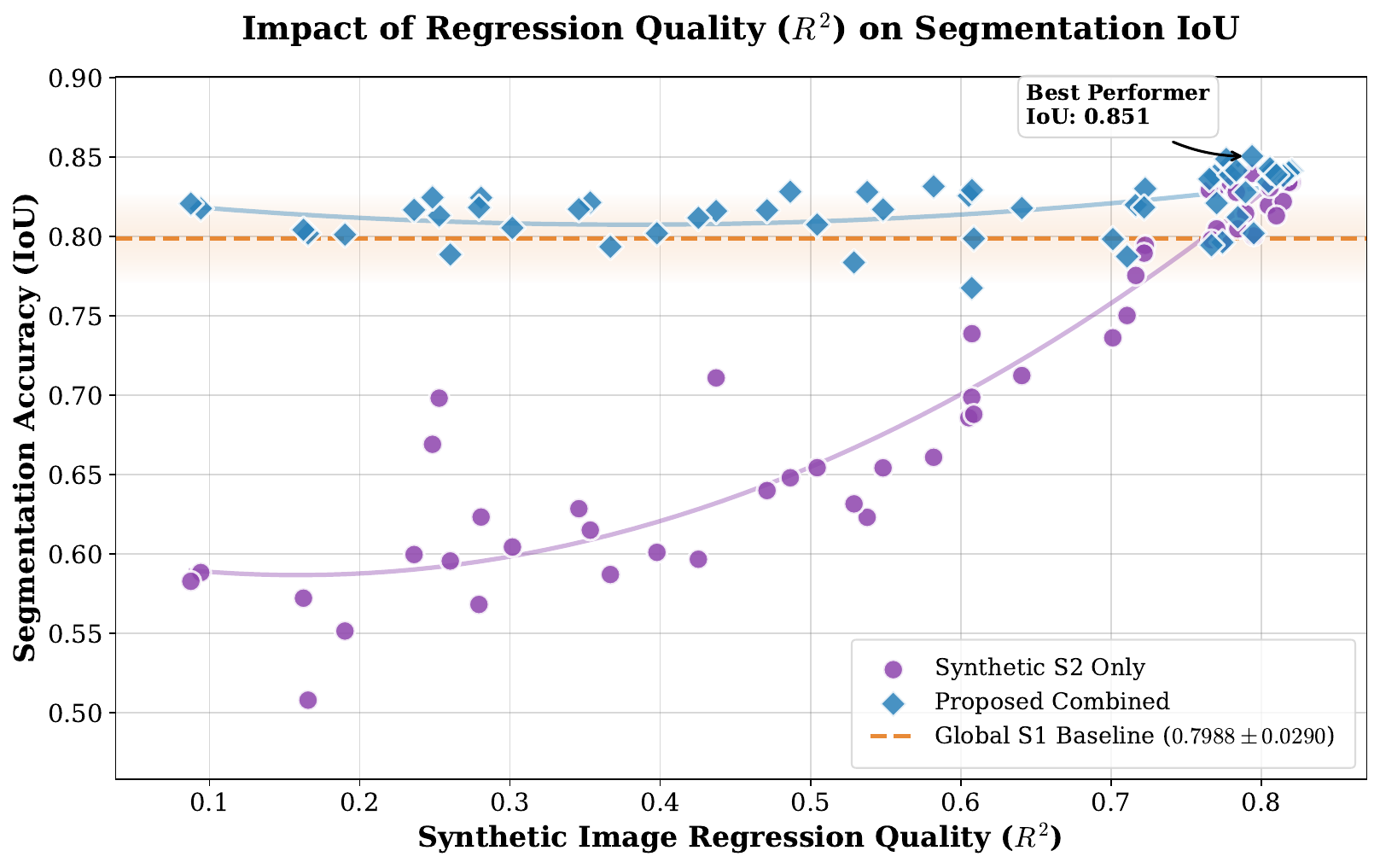}
    \caption{\textbf{Impact of regression quality on segmentation performance.} Scatter plot illustrating the relationship between the synthetic image regression quality ($R^2$) and the resulting segmentation accuracy (IoU). The \textcolor{purple}{purple circles} represent the Synthetic S2-only framework, while the \textcolor{blue}{blue diamonds} represent the Proposed Combined model. Solid trendlines indicate the polynomial fit for each approach. The horizontal \textcolor{orange}{dashed orange line} denotes the Global S1 Baseline ($\text{IoU} \approx .7988$), and the optimal combined configuration is explicitly annotated.}
    \label{fig:synthetic_s2_regression}
\end{figure}

\subsection{Impact of Synthetic Reconstruction Quality on Segmentation}
\label{sec:results_regression_quality}

To systematically evaluate how the fidelity of the generated optical features influences downstream task performance, we analyze the correlation between the synthetic NDWI regression quality ($R^2$) and the resulting segmentation accuracy (IoU). The distribution across various experimental folds and architectural capacities is illustrated in Fig.~\ref{fig:synthetic_s2_regression}. This experiment was focused on the mixed events evaluation because this allowed us to have a more uniform distribution as opposed to event-stratified evaluation where the geography is very skewed making this experiment very difficult to understand. As some locations might have very low variation with only dry land, this might make the comparison of $R^2$ very difficult; this is why the researchers focused more on the mixed events. 

A global baseline for standalone SAR performance is established at an average IoU of $0.7988 \pm 0.0290$ (indicated by the dashed orange line), derived from the mean performance of S1-only baseline experiments. Two primary empirical trends emerge from this analysis:

\begin{itemize}
    \item \textbf{Synthetic S2 Only:} Models relying exclusively on the generated synthetic NDWI exhibit a steep, non-linear positive correlation with regression quality. At low $R^2$ values ($< 0.40$), segmentation performance degrades severely (IoU dropping to between $0.50$ and $0.60$), falling substantially below the S1 baseline. However, as synthesis quality improves significantly ($R^2 > 0.75$), the Synthetic-only models consistently surpass the S1 baseline, reaching peak IoU scores around $0.84$. This highlights the high vulnerability of unimodal synthetic pipelines to upstream generative artifacts when regression quality is sub-optimal.
    
    \item \textbf{Proposed Combined Framework:} The multi-modal fusion approach demonstrates a significantly more robust and stable trajectory. Crucially, the Combined model maintains performance above the S1 baseline across nearly the entire spectrum of $R^2$ values, effectively mitigating the catastrophic failure modes observed in the Synthetic-only approach at low regression qualities. At the highest regression quality, the Combined framework achieves the global maximum segmentation IoU of $0.851$ (marked as the Best Performer). This confirms that integrating raw SAR features with synthetic optical features provides both baseline stabilization against poor generation and synergistic performance maximization when generation is accurate.
\end{itemize}

\begin{table*}[!ht]
\centering
\caption{Comprehensive ablation study on architectural capacity and input tile size under both Event-Stratified and Mixed cross-validation frameworks. Performance is evaluated using Intersection over Union (IoU) and F1-Score across Cross-Validation (CV) and Global Micro-aggregated metrics. The configuration is denoted as \textbf{Reg-[X] / Seg-[Y]}, indicating the capacity of the regression and segmentation backbones, respectively. Optimal metrics are \textbf{bolded}; second-best values are \underline{underlined}.}
\label{tab:ablation_architecture_tiles_iou_f1}
\renewcommand{\arraystretch}{1.0} 
\setlength{\tabcolsep}{6pt}
\small 
\begin{tabular}{lll cccc}
\toprule
\multirow{2}{*}{\textbf{Tile Size}} & \multirow{2}{*}{\textbf{Configuration}} & \multirow{2}{*}{\textbf{Input Modality}} & \multicolumn{2}{c}{\textbf{IoU}} & \multicolumn{2}{c}{\textbf{F1-Score}} \\
\cmidrule(lr){4-5} \cmidrule(lr){6-7}
& & & \textbf{CV (Mean $\pm$ Std)} & \textbf{Global} & \textbf{CV (Mean $\pm$ Std)} & \textbf{Global} \\
\midrule
\multicolumn{7}{c}{\textbf{Panel A: Event-Stratified Cross-Validation}} \\
\midrule
\multirow{12}{*}{$128 \times 128$} 
& \multirow{3}{*}{Reg-B0 / Seg-B0} 
  & S1 Only & \underline{0.6529 $\pm$ 0.1608} & \underline{0.7285} & \underline{0.7782 $\pm$ 0.1220} & \underline{0.8429} \\
& & Syn S2 Only & $0.5724 \pm 0.2633$ & $0.7197$ & $0.6847 \pm 0.2624$ & $0.8370$ \\
& & Combined & \textbf{0.6738 $\pm$ 0.1253} & \textbf{0.7503} & \textbf{0.7988 $\pm$ 0.0845} & \textbf{0.8574} \\
\cmidrule{2-7}
& \multirow{3}{*}{Reg-B3 / Seg-B0} 
  & S1 Only & $0.6150 \pm 0.1613$ & $0.7018$ & $0.7491 \pm 0.1254$ & $0.8248$ \\
& & Syn S2 Only & \underline{0.6203 $\pm$ 0.1479} & \underline{0.7187} & \underline{0.7557 $\pm$ 0.1093} & \underline{0.8364} \\
& & Combined & \textbf{0.6539 $\pm$ 0.1289} & \textbf{0.7291} & \textbf{0.7836 $\pm$ 0.0915} & \textbf{0.8433} \\
\cmidrule{2-7}
& \multirow{3}{*}{Reg-B3 / Seg-B3} 
  & S1 Only & \underline{0.6150 $\pm$ 0.1648} & $0.6813$ & \underline{0.7486 $\pm$ 0.1281} & $0.8105$ \\
& & Syn S2 Only & $0.6044 \pm 0.1599$ & \underline{0.7019} & $0.7417 \pm 0.1183$ & \underline{0.8248} \\
& & Combined & \textbf{0.6294 $\pm$ 0.1552} & \textbf{0.7099} & \textbf{0.7615 $\pm$ 0.1170} & \textbf{0.8304} \\
\cmidrule{2-7}
& \multirow{3}{*}{Reg-B5 / Seg-B0} 
  & S1 Only & $0.6334 \pm 0.1710$ & $0.7124$ & $0.7615 \pm 0.1352$ & $0.8320$ \\
& & Syn S2 Only & \underline{0.6559 $\pm$ 0.1395} & \underline{0.7316} & \underline{0.7840 $\pm$ 0.0975} & \underline{0.8450} \\
& & Combined & \textbf{0.6821 $\pm$ 0.1282} & \textbf{0.7577} & \textbf{0.8043 $\pm$ 0.0886} & \textbf{0.8622} \\

\midrule
\multirow{9}{*}{$256 \times 256$} 
& \multirow{3}{*}{Reg-B0 / Seg-B0} 
  & S1 Only & \textbf{0.4290 $\pm$ 0.0908} & $0.4354$ & \textbf{0.5949 $\pm$ 0.0859} & $0.6067$ \\
& & Syn S2 Only & \underline{0.4212 $\pm$ 0.2052} & \textbf{0.5063} & $0.5607 \pm 0.2224$ & \textbf{0.6722} \\
& & Combined & $0.4129 \pm 0.0983$ & \underline{0.4475} & \underline{0.5774 $\pm$ 0.1017} & \underline{0.6183} \\
\cmidrule{2-7}
& \multirow{3}{*}{Reg-B3 / Seg-B0} 
  & S1 Only & $0.4824 \pm 0.1901$ & $0.5542$ & $0.6311 \pm 0.1542$ & $0.7132$ \\
& & Syn S2 Only & \underline{0.5553 $\pm$ 0.2026} & \textbf{0.6610} & \underline{0.6937 $\pm$ 0.1579} & \textbf{0.7959} \\
& & Combined & \textbf{0.5619 $\pm$ 0.1982} & \underline{0.6583} & \textbf{0.7005 $\pm$ 0.1506} & \underline{0.7939} \\
\cmidrule{2-7}
& \multirow{3}{*}{Reg-B5 / Seg-B0} 
  & S1 Only & $0.5042 \pm 0.2071$ & $0.5818$ & $0.6483 \pm 0.1610$ & $0.7356$ \\
& & Syn S2 Only & \underline{0.5174 $\pm$ 0.2203} & \underline{0.6302} & \underline{0.6546 $\pm$ 0.1904} & \underline{0.7732} \\
& & Combined & \textbf{0.5596 $\pm$ 0.1946} & \textbf{0.6529} & \textbf{0.6988 $\pm$ 0.1521} & \textbf{0.7900} \\

\midrule
\multicolumn{7}{c}{\textbf{Panel B: Mixed Cross-Validation}} \\
\midrule
\multirow{12}{*}{$128 \times 128$} 
& \multirow{3}{*}{Reg-B0 / Seg-B0} 
  & S1 Only & $0.8068 \pm 0.0183$ & $0.8073$ & $0.8929 \pm 0.0114$ & $0.8934$ \\
& & Syn S2 Only & \underline{0.8162 $\pm$ 0.0138} & \underline{0.8168} & \underline{0.8987 $\pm$ 0.0084} & \underline{0.8992} \\
& & Combined & \textbf{0.8238 $\pm$ 0.0159} & \textbf{0.8243} & \textbf{0.9033 $\pm$ 0.0096} & \textbf{0.9037} \\
\cmidrule{2-7}
& \multirow{3}{*}{Reg-B3 / Seg-B0} 
  & S1 Only & $0.7988 \pm 0.0290$ & $0.8001$ & $0.8878 \pm 0.0182$ & $0.8890$ \\
& & Syn S2 Only & \underline{0.8236 $\pm$ 0.0147} & \underline{0.8241} & \underline{0.9032 $\pm$ 0.0089} & \underline{0.9036} \\
& & Combined & \textbf{0.8313 $\pm$ 0.0178} & \textbf{0.8318} & \textbf{0.9078 $\pm$ 0.0108} & \textbf{0.9082} \\
\cmidrule{2-7}
& \multirow{3}{*}{Reg-B3 / Seg-B3} 
  & S1 Only & $0.8163 \pm 0.0245$ & $0.8173$ & $0.8987 \pm 0.0150$ & $0.8994$ \\
& & Syn S2 Only & \underline{0.8259 $\pm$ 0.0129} & \underline{0.8265} & \underline{0.9046 $\pm$ 0.0078} & \underline{0.9050} \\
& & Combined & \textbf{0.8369 $\pm$ 0.0147} & \textbf{0.8375} & \textbf{0.9112 $\pm$ 0.0088} & \textbf{0.9115} \\
\cmidrule{2-7}
& \multirow{3}{*}{Reg-B5 / Seg-B0} 
  & S1 Only & $0.8049 \pm 0.0100$ & $0.8052$ & $0.8919 \pm 0.0061$ & $0.8921$ \\
& & Syn S2 Only & \underline{0.8279 $\pm$ 0.0154} & \underline{0.8285} & \underline{0.9058 $\pm$ 0.0093} & \underline{0.9062} \\
& & Combined & \textbf{0.8342 $\pm$ 0.0164} & \textbf{0.8347} & \textbf{0.9095 $\pm$ 0.0099} & \textbf{0.9099} \\

\midrule
\multirow{15}{*}{$256 \times 256$} 
& \multirow{3}{*}{Reg-B0 / Seg-B0} 
  & S1 Only & $0.7077 \pm 0.0260$ & $0.7097$ & $0.8286 \pm 0.0178$ & $0.8302$ \\
& & Syn S2 Only & \underline{0.7297 $\pm$ 0.0332} & \underline{0.7336} & \underline{0.8433 $\pm$ 0.0223} & \underline{0.8463} \\
& & Combined & \textbf{0.7373 $\pm$ 0.0315} & \textbf{0.7417} & \textbf{0.8484 $\pm$ 0.0212} & \textbf{0.8517} \\
\cmidrule{2-7}
& \multirow{3}{*}{Reg-B2 / Seg-B0} 
  & S1 Only & \underline{0.7390 $\pm$ 0.0000} & \underline{0.7390} & \underline{0.8499 $\pm$ 0.0000} & \underline{0.8499} \\
& & Syn S2 Only & $0.7241 \pm 0.0000$ & $0.7241$ & $0.8400 \pm 0.0000$ & $0.8400$ \\
& & Combined & \textbf{0.7422 $\pm$ 0.0000} & \textbf{0.7422} & \textbf{0.8520 $\pm$ 0.0000} & \textbf{0.8520} \\
\cmidrule{2-7}
& \multirow{3}{*}{Reg-B3 / Seg-B0} 
  & S1 Only & $0.7278 \pm 0.0380$ & $0.7302$ & $0.8419 \pm 0.0255$ & $0.8441$ \\
& & Syn S2 Only & \underline{0.7506 $\pm$ 0.0555} & \underline{0.7570} & \underline{0.8564 $\pm$ 0.0365} & \underline{0.8617} \\
& & Combined & \textbf{0.7630 $\pm$ 0.0531} & \textbf{0.7694} & \textbf{0.8645 $\pm$ 0.0347} & \textbf{0.8697} \\
\cmidrule{2-7}
& \multirow{3}{*}{Reg-B3 / Seg-B3} 
  & S1 Only & $0.7307 \pm 0.0290$ & $0.7329$ & $0.8441 \pm 0.0193$ & $0.8459$ \\
& & Syn S2 Only & \underline{0.7363 $\pm$ 0.0532} & \underline{0.7411} & \underline{0.8470 $\pm$ 0.0356} & \underline{0.8513} \\
& & Combined & \textbf{0.7563 $\pm$ 0.0427} & \textbf{0.7603} & \textbf{0.8606 $\pm$ 0.0279} & \textbf{0.8638} \\
\cmidrule{2-7}
& \multirow{3}{*}{Reg-B5 / Seg-B0} 
  & S1 Only & $0.7230 \pm 0.0272$ & $0.7247$ & $0.8389 \pm 0.0183$ & $0.8404$ \\
& & Syn S2 Only & \underline{0.7431 $\pm$ 0.0469} & \underline{0.7470} & \underline{0.8518 $\pm$ 0.0308} & \underline{0.8552} \\
& & Combined & \textbf{0.7586 $\pm$ 0.0358} & \textbf{0.7615} & \textbf{0.8623 $\pm$ 0.0230} & \textbf{0.8646} \\

\bottomrule
\end{tabular}
\end{table*}

\begin{table}[!ht]
\centering
\caption{Statistical significance testing (Wilcoxon signed-rank test) for image-wise Intersection over Union (IoU). P-values indicate whether the performance improvements of alternative input modalities over baselines are significant ($p < 0.05$, marked in \textbf{bold}).}
\label{tab:ablation_stats_iou_global}
\renewcommand{\arraystretch}{1.1}
\setlength{\tabcolsep}{6pt}
\small
\begin{tabular}{ll ccc}
\toprule
\multirow{2}{*}{\textbf{Tile Size}} & \multirow{2}{*}{\textbf{Configuration}} & \multicolumn{3}{c}{\textbf{Global Scenes (p-values)}} \\
\cmidrule(lr){3-5}
& & \textbf{Comb $>$ S1} & \textbf{Syn S2 $>$ S1} & \textbf{Comb $>$ Syn S2} \\
\midrule
\multicolumn{5}{c}{\textbf{Panel A: Event-Stratified Cross-Validation}} \\
\midrule
\multirow{4}{*}{$128 \times 128$} & Reg-B0 / Seg-B0 & $\mathbf{1.47} \times \mathbf{10}^{\mathbf{-78}}$ & $1.000$ & $\mathbf{3.32} \times \mathbf{10}^{\mathbf{-149}}$ \\
& Reg-B3 / Seg-B0 & $\mathbf{1.88} \times \mathbf{10}^{\mathbf{-73}}$ & $\mathbf{5.69} \times \mathbf{10}^{\mathbf{-13}}$ & $\mathbf{8.78} \times \mathbf{10}^{\mathbf{-27}}$ \\
& Reg-B3 / Seg-B3 & $1.000$ & $1.000$ & $\mathbf{2.43} \times \mathbf{10}^{\mathbf{-43}}$ \\
& Reg-B5 / Seg-B0 & $\mathbf{1.49} \times \mathbf{10}^{\mathbf{-113}}$ & $\mathbf{3.15} \times \mathbf{10}^{\mathbf{-17}}$ & $\mathbf{1.11} \times \mathbf{10}^{\mathbf{-35}}$ \\

\cmidrule{2-5}
\multirow{3}{*}{$256 \times 256$} & Reg-B0 / Seg-B0 & $\mathbf{7.34} \times \mathbf{10}^{\mathbf{-12}}$ & $0.853$ & $\mathbf{5.28} \times \mathbf{10}^{\mathbf{-22}}$ \\
& Reg-B3 / Seg-B0 & $\mathbf{6.41} \times \mathbf{10}^{\mathbf{-19}}$ & $\mathbf{0.013}$ & $\mathbf{2.73} \times \mathbf{10}^{\mathbf{-9}}$ \\
& Reg-B5 / Seg-B0 & $\mathbf{5.39} \times \mathbf{10}^{\mathbf{-5}}$ & $0.260$ & $\mathbf{2.53} \times \mathbf{10}^{\mathbf{-3}}$ \\

\midrule
\multicolumn{5}{c}{\textbf{Panel B: Mixed Cross-Validation}} \\
\midrule
\multirow{4}{*}{$128 \times 128$} & Reg-B0 / Seg-B0 & $\mathbf{1.87} \times \mathbf{10}^{\mathbf{-45}}$ & $0.642$ & $\mathbf{3.39} \times \mathbf{10}^{\mathbf{-82}}$ \\
& Reg-B3 / Seg-B0 & $\mathbf{5.64} \times \mathbf{10}^{\mathbf{-68}}$ & $\mathbf{2.09} \times \mathbf{10}^{\mathbf{-15}}$ & $\mathbf{2.08} \times \mathbf{10}^{\mathbf{-47}}$ \\
& Reg-B3 / Seg-B3 & $\mathbf{2.52} \times \mathbf{10}^{\mathbf{-19}}$ & $0.927$ & $\mathbf{2.99} \times \mathbf{10}^{\mathbf{-54}}$ \\
& Reg-B5 / Seg-B0 & $\mathbf{5.80} \times \mathbf{10}^{\mathbf{-71}}$ & $\mathbf{1.36} \times \mathbf{10}^{\mathbf{-18}}$ & $\mathbf{1.52} \times \mathbf{10}^{\mathbf{-38}}$ \\

\cmidrule{2-5}
\multirow{5}{*}{$256 \times 256$} & Reg-B0 / Seg-B0 & $\mathbf{1.18} \times \mathbf{10}^{\mathbf{-25}}$ & $\mathbf{6.69} \times \mathbf{10}^{\mathbf{-12}}$ & $\mathbf{1.84} \times \mathbf{10}^{\mathbf{-9}}$ \\
& Reg-B2 / Seg-B0 & $\mathbf{2.27} \times \mathbf{10}^{\mathbf{-5}}$ & $\mathbf{1.57} \times \mathbf{10}^{\mathbf{-6}}$ & $0.999$ \\
& Reg-B3 / Seg-B0 & $\mathbf{8.42} \times \mathbf{10}^{\mathbf{-37}}$ & $\mathbf{3.16} \times \mathbf{10}^{\mathbf{-19}}$ & $\mathbf{7.24} \times \mathbf{10}^{\mathbf{-13}}$ \\
& Reg-B3 / Seg-B3 & $\mathbf{3.33} \times \mathbf{10}^{\mathbf{-35}}$ & $\mathbf{2.00} \times \mathbf{10}^{\mathbf{-23}}$ & $\mathbf{0.031}$ \\
& Reg-B5 / Seg-B0 & $\mathbf{2.53} \times \mathbf{10}^{\mathbf{-48}}$ & $\mathbf{3.99} \times \mathbf{10}^{\mathbf{-36}}$ & $\mathbf{2.96} \times \mathbf{10}^{\mathbf{-9}}$ \\

\bottomrule
\end{tabular}
\end{table}

\subsection{Ablation Studies on Model Capacity and Tile Size}
\label{subsec:ablations}

To systematically evaluate the impact of spatial tile dimensions, architectural backbone capacity, and input modalities, a multi-dimensional ablation study was conducted. Performance was assessed across both Event-Stratified and Mixed cross-validation regimes on standard ($128 \times 128$) and extended ($256 \times 256$) tile sizes. Primary performance metrics—Intersection over Union (IoU) and F1-Score—are presented in Table~\ref{tab:ablation_architecture_tiles_iou_f1}, while formal statistical significance testing using image-wise Wilcoxon signed-rank tests is reported in Table~\ref{tab:ablation_stats_iou_global}. To provide a holistic view of model behavior, supplementary metrics including Specificity and Recall (Table~\ref{tab:ablation_architecture_tiles_spec_recall}) as well as Precision and Accuracy (Table~\ref{tab:ablation_architecture_tiles_prec_acc}) are integrated into this analysis.

\vspace{0.1cm}
\noindent\textbf{Validation Framework Dynamics: Event-Stratified vs. Mixed CV:}
A stark performance disparity is observed between the two cross-validation frameworks. Under Mixed Cross-Validation, where geographical scenes are randomly partitioned across folds, models achieve higher baseline performance due to spatial feature familiarity (e.g., Global IoU reaching up to $0.8375$). Conversely, Event-Stratified Cross-Validation forces zero-shot generalization across unseen flood disaster events, introducing substantial distribution shifts. In this challenging regime, raw performance metrics contract (e.g., Global IoU ranging from $0.4354$ to $0.7577$), highlighting the necessity of multi-modal generative fusion to mitigate spatial variance.

\vspace{0.1cm}
\noindent\textbf{Standard Context Performance ($128 \times 128$):}
At the standard $128 \times 128$ resolution, the Combined input modality consistently outperforms both single-modality baselines (S1 Only and Synthetic S2 Only) across nearly all backbone configurations:
\begin{itemize}
    \item \textbf{Event-Stratified Regime:} Even with a lightweight backbone (Reg-B0 / Seg-B0), the Combined framework achieves a robust Global IoU of $0.7503$ and F1-Score of $0.8574$, significantly outpacing the S1 Only baseline ($0.7285$ IoU). Scaling the regression encoder capacity to MiT-B5 (Reg-B5 / Seg-B0) yields the peak performance at this tile resolution, reaching a CV IoU of $0.6821 \pm 0.1282$, a Global IoU of $0.7577$, and a Global F1-Score of $0.8622$ (Table~\ref{tab:ablation_architecture_tiles_iou_f1}).
    \item \textbf{Trade-off Analysis (Precision vs. Recall):} Analysis of supplementary metrics reveals distinct modality behaviors. As shown in Table~\ref{tab:ablation_architecture_tiles_spec_recall} and Table~\ref{tab:ablation_architecture_tiles_prec_acc}, Synthetic S2 Only models consistently yield superior Specificity (e.g., up to $0.9885$ in Event-Stratified Reg-B3 / Seg-B3) and Precision (e.g., up to $0.8302$ in Event-Stratified Reg-B3 / Seg-B0) by producing fewer false positives in land cover regions. However, the Combined framework maximizes Recall (e.g., reaching $0.7972 \pm 0.1108$ CV Recall and $0.8574$ Global Recall for Reg-B5 / Seg-B0) and overall Accuracy ($0.9724$), proving that combining SAR and synthetic optical modalities suppresses false negatives during complex inundation scenarios.
\end{itemize}

\vspace{0.1cm}
\noindent\textbf{Model Overparameterization and Training Stability:}
While scaling generally improves performance, deploying high-capacity segmentation decoders introduces severe diminishing returns and practical challenges. Under Mixed CV, expanding both regression and segmentation encoders to MiT-B3 (Reg-B3 / Seg-B3) achieves the highest overall accuracy, attaining a CV IoU of $0.8369 \pm 0.0147$ and a Global IoU of $0.8375$ (Table~\ref{tab:ablation_architecture_tiles_prec_acc}). However, empirical observations during optimization reveal that the Seg-B3 configuration is distinctly overparameterized for this specific flood segmentation task. Compared to the robust and smooth convergence of the lightweight Seg-B0 configurations, the training dynamics of the Seg-B3 models were highly unstable. This instability suggests that while heavier segmentation heads can memorize features in easier Mixed regimes, the simpler Seg-B0 heads are vastly more reliable and sufficient when paired with a highly capable regression encoder.

\vspace{0.1cm}
\noindent\textbf{The Spatial Capacity Bottleneck ($256 \times 256$):}
A critical limitation emerges when input tile dimensions are quadrupled to $256 \times 256$ without sufficient model parameter depth. On $256 \times 256$ inputs under Event-Stratified CV:
\begin{itemize}
    \item The compact Combined Reg-B0 / Seg-B0 model experiences severe performance degradation, with its CV IoU falling to $0.4129 \pm 0.0983$ and Global IoU dropping to $0.4475$. Here, the Combined setup actually underperforms the Synthetic S2 Only baseline, which achieves a Global IoU of $0.5063$ and Global F1-Score of $0.6722$ (Table~\ref{tab:ablation_architecture_tiles_iou_f1}).
    \item Supplementary analysis reveals that this bottleneck is primarily driven by a collapse in Recall ($0.5925 \pm 0.2377$ for Combined vs. $0.6858$ Global Recall for S1 Only, Table~\ref{tab:ablation_architecture_tiles_spec_recall}) and Precision ($0.5712$ Global Precision, Table~\ref{tab:ablation_architecture_tiles_prec_acc}). Lightweight encoders lack the receptive field depth and parameter capacity needed to map complex SAR-to-NDWI translations across expanded spatial contexts, thereby propagating generative artifacts that distort downstream segmentation masks.
\end{itemize}

\vspace{0.1cm}
\noindent\textbf{Resolving Spatial Bottlenecks via Architectural Scaling:}
Scaling the generative backbone capacity resolves this bottleneck on $256 \times 256$ tiles by providing the expressive capacity necessary to process wider spatial contexts, avoiding the need for an overparameterized segmentation head:
\begin{itemize}
    \item Upgrading to the Reg-B3 / Seg-B0 configuration in Event-Stratified CV restores multi-modal superiority, raising CV IoU to $0.5619 \pm 0.1982$, Global IoU to $0.6583$, and Global F1-Score to $0.7939$ (Table~\ref{tab:ablation_architecture_tiles_iou_f1}).
    \item Under Mixed CV at $256 \times 256$, the Reg-B3 / Seg-B0 Combined model attains peak global metrics, leading with a Global IoU of $0.7694$, a Global F1-Score of $0.8697$, a Global Recall of $0.8323$, and a Global Precision of $0.9106$. This confirms that high-capacity regression backbones effectively utilize larger context windows to filter noise and refine target boundaries.
\end{itemize}

\vspace{0.1cm}
\noindent\textbf{Statistical Significance Analysis:}
To verify that the observed performance gains of the Combined modality are not artifacts of sampling variance, paired image-wise Wilcoxon signed-rank tests were performed on global IoU distributions ($p < 0.05$ threshold, Table~\ref{tab:ablation_stats_iou_global}):
\begin{itemize}
    \item \textbf{Combined vs. Single Baselines:} In almost all configurations, the improvement of Combined over S1 Only ($\text{Comb} > \text{S1}$) is statistically significant with extremely small p-values (e.g., $p = 1.47 \times 10^{-78}$ for $128 \times 128$ Reg-B0 / Seg-B0 in Panel A, and $p = 1.18 \times 10^{-25}$ for $256 \times 256$ Reg-B0 / Seg-B0 in Panel B). Similarly, the superiority of Combined over Synthetic S2 Only ($\text{Comb} > \text{Syn S2}$) is strongly supported (e.g., $p = 3.32 \times 10^{-149}$ for $128 \times 128$ Reg-B0 / Seg-B0 in Panel A).
    \item \textbf{Boundary Cases:} The statistical tests also mathematically confirm architectural bottlenecks. For example, in Event-Stratified $128 \times 128$ Reg-B3 / Seg-B3, the $\text{Comb} > \text{S1}$ comparison yields $p = 1.000$ (non-significant), aligning with the marginal CV IoU difference ($0.6294$ vs. $0.6150$). Likewise, in Mixed $256 \times 256$ Reg-B2 / Seg-B0, $\text{Comb} > \text{Syn S2}$ yields $p = 0.999$, reflecting identical Global Precision ($0.9047$ vs. $0.9022$) and near-identical IoU metrics.
\end{itemize}

In summary, these ablation experiments establish a key architectural guideline for multi-modal generative segmentation: while multi-modal fusion reliably improves flood detection accuracy, maximum stability and performance require scaling the generative module's capacity proportionally with input spatial dimensions while keeping the downstream segmentation head lightweight to prevent overparameterization.

\begin{figure}
    \centering
    \includegraphics[width=0.9\linewidth]{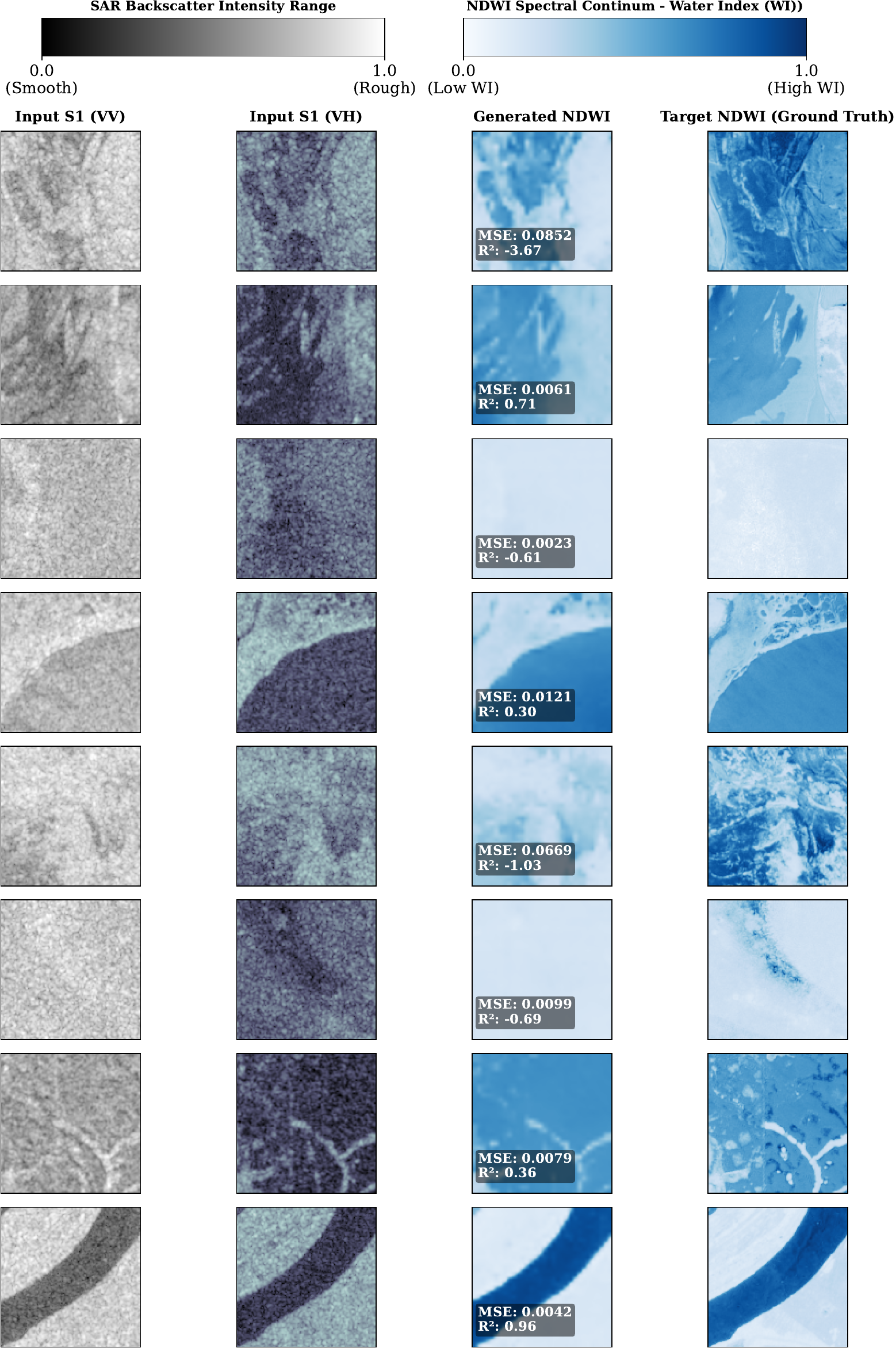}
    
\caption{\textbf{Qualitative evaluation of synthetic NDWI generation.} The top row displays the colorbar scales for the input Sentinel-1 SAR data (left) and the Normalized Difference Water Index (NDWI) continuum (right). The image rows track representative test patches, demonstrating how radar backscatter maps to continuous water index values. The columns show the Input SAR (first and second column), Predicted NDWI (third column), and Target NDWI (fourth column). Bounding boxes display patch-level Mean Squared Error (MSE) and Coefficient of Determination ($R^2$). This high-fidelity translation provides the essential spectral mapping required to guide the downstream binary water segmentation task.}

    \label{fig:ndwi_generation}
\end{figure}

\begin{figure*}[!htbp]
\begin{center}
   
    \includegraphics[width=\linewidth]{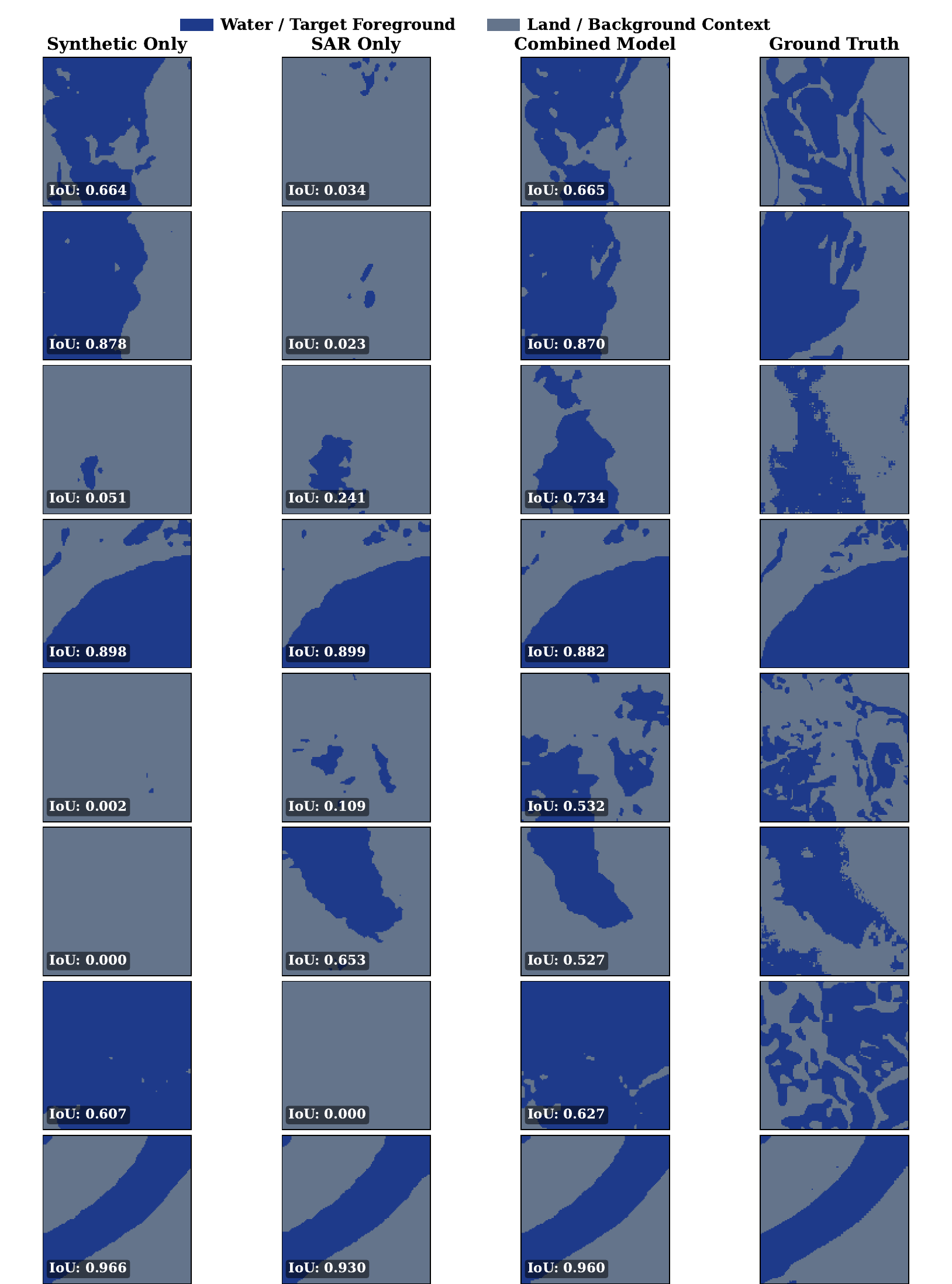}
    \caption{\textbf{Comparative analysis of downstream water segmentation performance.} The top row provides a legend defining the binary classification mapping for target surface water (dark blue) and terrestrial land background context (slate gray). The image matrix evaluates model segmentation performance across representative test patches. The columns show predictions from the Synthetic-Only model (first column), raw SAR-Only model (second column), Combined translation-segmentation model (third column), and the Ground Truth target mask (fourth column). Bounding boxes display patch-level Intersection over Union (IoU) metrics, illustrating how integrating synthetic index translation mitigates backscatter noise and stabilizes surface feature extraction.}
    \label{fig:downstream_segmentation}
\end{center}
\end{figure*}

\subsection{Qualitative Analysis and Dual-Stage Performance}
\label{subsec:qualitative_analysis}

A visual inspection of the image patches confirms our quantitative findings and demonstrates how the framework handles diverse landscapes. Figure~\ref{fig:ndwi_generation} illustrates the continuous NDWI generation stage, while Figure~\ref{fig:downstream_segmentation} evaluates the final binary water segmentation stage. Because both figures analyze the exact same geographical patches in a row-by-row sequence, we can directly observe how the quality of the intermediate translation shapes the final water mask.

\subsubsection{Continuous Translation Evaluation (Figure~\ref{fig:ndwi_generation})}
The continuous regression module demonstrates high pixel-level accuracy by maintaining low Mean Squared Error (MSE) values across all test scenes. 

\begin{itemize}
    \item \textbf{High-Contrast Features (Rows 2, 4, 7, and 8):} In scenes containing distinct coastlines or river networks, the model accurately reconstructs the underlying geometry, achieving high localized $R^2$ scores up to $0.96$. 
    \item \textbf{Uniform Landscapes (Rows 1, 3, 5, and 6):} In uniform patches consisting almost entirely of land, the localized $R^2$ drops into negative values (down to $-3.67$). Crucially, a negative $R^2$ in this context is not an indication of poor model performance. Because these patches contain almost zero spatial or pixel variance, the mathematical denominator of the $R^2$ metric approaches zero, making it hyper-sensitive to minor deviations that the model cannot perfectly match. Because these negative $R^2$ values are paired with near-zero errors (e.g., $\text{MSE} = 0.0023$ in Row 3), the visuals and metrics combined confirm that the model successfully matches the true absolute numeric index values of the terrain.
\end{itemize}

\subsubsection{Downstream Water Mask Segmentation (Figure~\ref{fig:downstream_segmentation})}
Evaluating the binary masks reveals a robust, bidirectional error-correction mechanism within the proposed multi-view framework:

\begin{itemize}
    \item \textbf{Synthetic Branch Rectifies Specular SAR Failure (Rows 1, 2, 3, and 7):} In these scenes, smooth terrain or specular water reflections cause the standalone \textit{SAR Only} model to misclassify surface water, resulting in failed segmentations ($\text{IoU} \leq 0.034$). However, the \textit{Synthetic Only} branch successfully builds the correct index layouts. By utilizing these synthetic features, the \textit{Combined Model} overrides the corrupted radar backscatter, restoring high-accuracy masks ($\text{IoU} = 0.665$, $0.870$, $0.734$, and $0.627$).
    
    \item \textbf{Raw SAR Rectifies Generative Drop-Out (Rows 5 and 6):} Conversely, where low-contrast boundaries or complex land textures cause the \textit{Synthetic Only} branch to miss water bodies entirely ($\text{IoU} = 0.000$), the physical surface roughness caught by the radar sensor acts as a backup. The \textit{SAR Only} model successfully detects these features ($\text{IoU} = 0.109$ and $0.653$). The \textit{Combined Model} leverages this structural anchor to recover the hidden geometry, stabilizing performance with IoU scores of $0.532$ and $0.527$.
    
    \item \textbf{Mutual Reinforcement (Rows 4 and 8):} When a patch contains clean, uncorrupted features—such as the massive coastline in Row 4 or the broad river channel in Row 8—both independent streams perform optimally. The \textit{Combined Model} capitalizes on this shared clarity to output near-perfect geometric boundaries ($\text{IoU} = 0.882$ and $0.960$).
\end{itemize}

This dual-stage analysis proves that while isolated data views remain vulnerable to environmental anomalies, the combined multi-view network reliably self-corrects to maintain stable, cloud-resilient surface water monitoring.
\section{Discussion}
\label{sec:discussion}

The empirical findings presented in Section~\ref{sec:results} validate that fusing synthetic optical representations with raw SAR data significantly enhances hydrological monitoring under cloud cover. These results reveal critical insights into modality synergies, non-linear feature extraction, and structural error correction.

\subsection{Modality Synergies and Non-Linear Feature Extraction}
\label{sec:discussion_multi_view_fusion}

Our experiments establish that the translation of SAR to synthetic optical (NDWI) features represents a profound structural clean-up of the input data rather than a simple redistribution of statistical variance. Under high-fidelity conditions, a segmentation network utilizing purely synthetic S2 data outperforms the original raw S1 baseline (e.g., achieving a Global IoU of 0.8285 versus 0.8052 under Mixed splitting, Table~\ref{table:main_results}). 

To isolate whether this performance gain stems from deep semantic feature extraction or mere dimensionality reduction, we benchmarked the generative approach against a Principal Component Analysis (S1 PCA) baseline. The analysis yields two critical observations:
\begin{itemize}
    \item \textbf{Generative Translation vs. Linear Transformation:} While applying PCA to raw S1 images provides a minor performance lift in Event-Stratified regimes (improving Global IoU from 0.7124 to 0.7263), it fails completely in Mixed regimes, yielding no statistically significant improvement over raw S1 data ($p = 0.9999$, Table~\ref{table:main_results} Panel D). In contrast, the deep regression network actively filters out domain-specific SAR noise—such as speckle and layover artifacts—mapping the chaotic radar backscatter into a clean, highly separable optical proxy space.
    \item \textbf{Statistical Validation of Semantic Clean-up:} The superiority of the deep regression model over linear PCA is mathematically proven across all validation frameworks. Wilcoxon signed-rank tests confirm that the Synthetic S2 model significantly outperforms the S1 PCA baseline in both Event-Stratified ($p = 1.62 \times 10^{-3}$) and Mixed ($p = 1.51 \times 10^{-29}$) regimes (Table~\ref{table:main_results}).
\end{itemize}

However, relying solely on synthetic data introduces operational risks if the generator encounters sub-optimal conditions and hallucinates features. The superior stability of the Combined (Synth S2 + S1) framework—which achieves the highest Global IoU of 0.7577 (Event-Stratified) and 0.8347 (Mixed)—demonstrates the necessity of multi-view learning:
\begin{itemize}
    \item \textbf{Structural Anchoring:} When the synthetic generator produces noisy or inaccurate features, the Combined network avoids collapse by dynamically falling back on raw S1 backscatter, using the physical SAR data as a structural safety net.
    \item \textbf{Semantic Augmentation:} At the upper bound of generator quality, the network effectively fuses the geometric texture of raw SAR with the clean boundaries of the synthetic optical data. This multi-modal fusion consistently produces statistically significant improvements over both the S1 PCA and S1 PCA + Raw S1 baselines ($p < 0.0001$ across all panels).
\end{itemize}

\subsection{Model Capacity and Spatial Scaling Trade-offs}
\label{sec:discussion_scaling_laws}

Our cross-validation analysis shows that scaling the generative regression module yields critical improvements for downstream segmentation. High-capacity regression backbones (e.g., MiT-B3 or B5) capture the subtle semantic nuances necessary for resolving complex, irregular boundaries along coastlines and riparian zones, effectively filtering noise over wider context windows.

This capacity requirement becomes acute when scaling to larger spatial contexts ($256 \times 256$). Our analysis reveals a distinct spatial capacity bottleneck: lightweight configurations lack the parameter depth to maintain spatial coherence over wider contexts, thereby introducing generative errors. 

Crucially, however, scaling both halves of the architecture is counterproductive. Expanding the segmentation head to match the regression backbone introduces severe overparameterization and training instability for the segmentation task. Our findings indicate that pairing a high-capacity regression encoder with a lightweight segmentation decoder offers the optimal practical balance. This configuration successfully captures broader contextual semantics without suffering from optimization collapse, securing highly stable and accurate multi-modal predictions.

\subsection{Failure Modes and Bidirectional Rectification}
\label{sec:discussion_failures}

Qualitative analysis highlights how the proposed framework achieves bidirectional error correction under challenging environmental conditions:
\begin{itemize}
    \item \textbf{SAR Failure Rectification:} Standalone SAR models frequently fail when calm, smooth water surfaces act as specular reflectors. This scatters radar energy away from the sensor, creating backscatter signatures that mimic dry soil. In these scenarios, the Synthetic Branch successfully reconstructs the expected continuous NDWI profile by leveraging learned contextual cues, bypassing the physical limitations of the radar sensor.
    \item \textbf{Generative Failure Rectification:} Conversely, when the regression generator produces localized artifacts due to complex or unseen land cover textures, the Combined framework suppresses these hallucinations by forcing a reliance on the raw SAR structural features. 
\end{itemize}
This bidirectional correction loop prevents catastrophic segmentation failures and ensures generalizable performance across diverse and challenging hydrological conditions.
\section{Conclusion}
\label{sec:conclusion}

This study demonstrates that, under complete cloud cover, combining raw radar data with translated synthetic optical images using a scaled, pre-trained model yields optimal performance for water body segmentation. Our findings reveal four core principles governing this multi-modal pipeline. First, a direct positive correlation exists between downstream segmentation accuracy and the quality ($R^2$) of the intermediate synthetic reconstruction, proving that the translation process acts as a semantic filter where maximizing optical image quality directly enhances final mapping results. Second, for this optical index generation task, leveraging transfer learning on pre-trained models proves to be significantly more accurate and parameter-efficient than utilizing architectures trained from scratch. Third, the fusion of both modalities establishes a robust safety net where each stream counteracts the vulnerabilities of the other; the synthetic stream effectively corrects radar-specific errors like specular reflective glare, while physical radar data maintains geometric stability during sudden generative failures. Finally, a strict capacity scaling law dictates that larger spatial footprints introduce a structural bottleneck, meaning that model capacity must scale proportionally with the image window to prevent generative artifacts. Together, these interconnected principles provide a clear, reliable blueprint for building stable, all-weather environmental monitoring systems.


\section{Declarations}
\subsection*{Competing interests}
The authors declare that there are no competing interests.

\subsection*{Funding}
This research work was supported by a RISE Internal Research Grant with ID 2022-01-025. Nowreen and Rahman are partially supported by Basic Research Grants of BUET.

\subsection*{Use of AI}
ChatGPT-3.5 (OpenAI) was used solely for grammatical correction and readability improvements.

\section{Code and Data Availability}

All experiments were conducted using Colab Pro with a T4 GPU, 51 GB of system RAM, and 15 GB of GPU RAM. The complete code is available at:  

\url{https://github.com/bojack-horseman91/NDWI-generator}

The dataset used in this study is available at:

\url{https://cmr.earthdata.nasa.gov/search/concepts/C2781412798-MLHUB.html}

\FloatBarrier

\bibliography{sn-bibliography}
\backmatter
\fontfamily{cmr}\selectfont 
\setcounter{section}{0}
\setcounter{figure}{0}
\setcounter{table}{0}

\renewcommand{\thesection}{S\arabic{section}}
\renewcommand{\thefigure}{S\arabic{figure}}
\renewcommand{\thetable}{S\arabic{table}}

\section*{S1. Detailed Analysis of Pretraining and Model Architecture}

\begin{figure*}[!htbp]
    \centering
   
    \includegraphics[width=0.95\linewidth]{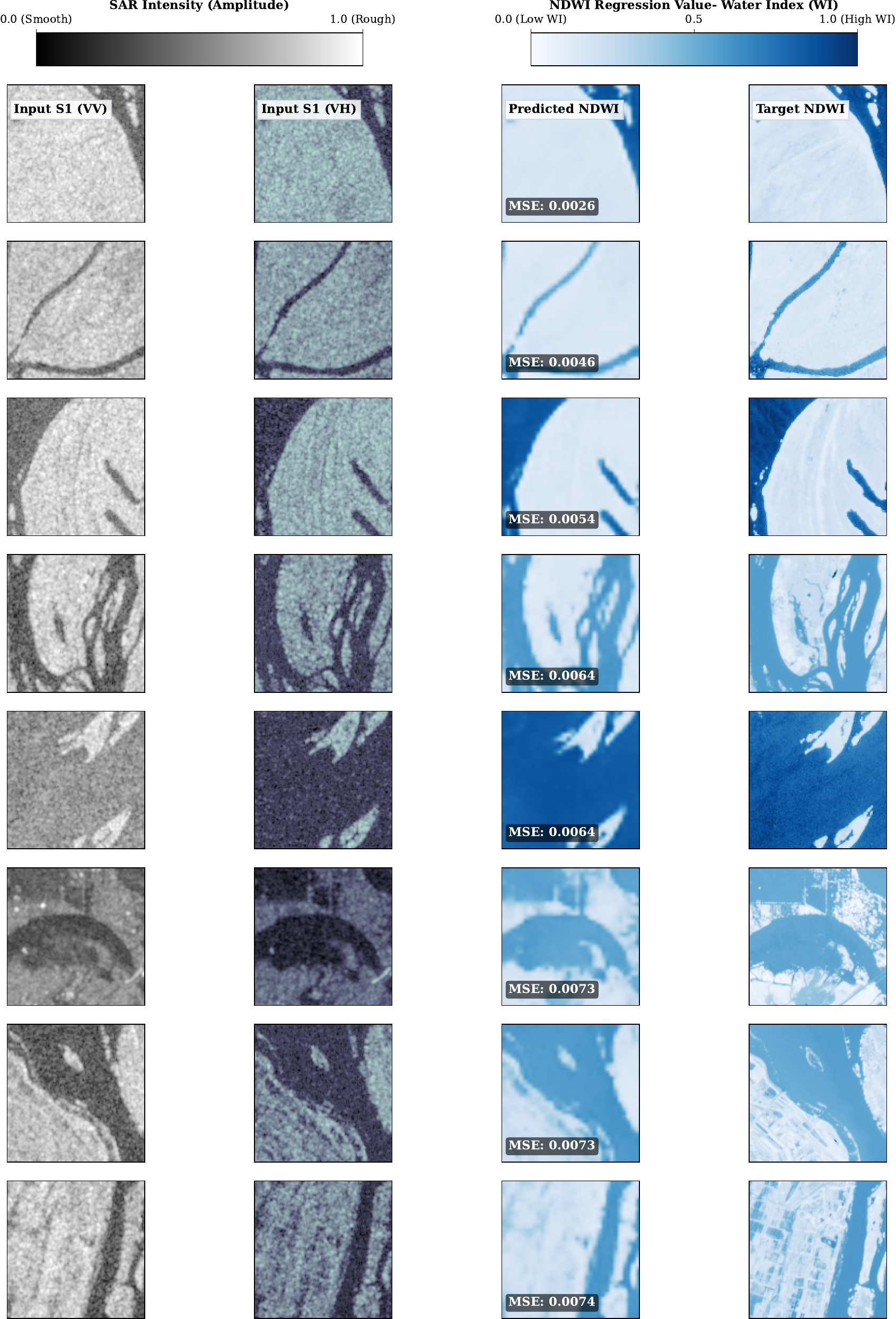}
    \caption{\textbf{High-performance NDWI predictions from SAR data.} These examples display the model's ability to synthesize optical water indices directly from Sentinel-1 inputs. The columns show the Input SAR (first and second column), Target NDWI (third column), and Predicted NDWI (fourth column). The consistently low MSE values ($0.0026-0.0074$) and the visual preservation of water boundaries highlight the model's capacity to suppress speckle noise while recovering structural details.}
    \label{fig:ndwi_predictions}
\end{figure*}
\label{sec:supp_pretraining}

As introduced in the main manuscript, a comparative analysis was conducted to evaluate the efficacy of transfer learning versus training from scratch for the generation of Synthetic NDWI. The experiments compared established architectures (e.g., U-Net, ViT, Swin Transformer) initialized with random weights against the SegFormer family utilizing encoders pretrained on ImageNet. 

\subsection*{S1.1. Superiority of Pretrained Transformer Architectures}
As evidenced by the quantitative results (Table \ref{table:performance} in the main text), the SegFormer variants utilizing ImageNet-pretrained encoders consistently outperform the custom architectures trained from scratch across all metrics. The approximately $15\%$ performance uplift observed in the SegFormer variants indicates that the synergistic combination of advanced attention mechanisms and learned semantic weights provides a vastly superior optimization landscape. While it is acknowledged that this comparison encompasses both a shift in architecture (CNN to Transformer) and initialization strategy (scratch to pretrained), for the empirical scope of the proposed framework, pretrained Transformer encoders represent the definitive optimal baseline for high-fidelity cross-modal regression.

\subsection*{S1.2. Domain Adaptation and Generalization}
Crucially, these results highlight the effectiveness of transfer learning even across disparate domains. Although the SegFormer encoders were pretrained on ImageNet (comprising natural images), the learned hierarchical features—ranging from low-level edges and textures to high-level semantic structures—prove highly transferable to the remote sensing domain. Remote sensing datasets are often constrained in size compared to massive computer vision benchmarks. By initializing the network with generalized weights, the model avoids overfitting to the specific speckle noise patterns of the SAR training data. Instead, it fine-tunes robust feature extractors that generalize better to unseen satellite imagery.

\subsection*{S1.3. Parameter Efficiency and Feature Quality}
A critical observation from the comparative analysis is the inverse relationship between parameter count and performance when comparing models trained from scratch versus pretrained models. The largest U-Net variant evaluated ($124.39$M parameters) yielded a lower $R^2$ ($0.6031$) than the smallest SegFormer-b0 variant ($3.70$M parameters, $R^2$ $0.7363$). This indicates that raw model capacity alone is insufficient for high-fidelity regression; the quality of the initialized learned features is paramount. The pretrained b0 model achieves superior results with roughly $3\%$ of the parameter count of the large U-Net, highlighting extreme computational efficiency and reinforcing the necessity of transfer learning for the efficient synthesis of optical indices from SAR data.

\begin{table*}[!ht]
\centering
\caption{\textbf{Supplementary:} Extended performance metrics (Accuracy, Precision, Recall, and Specificity) separated into Cross-Validation and Global results. The segmentation network utilizes SegFormer (MiT-B0), and the synthetic representations are generated using the optimal MiT-B5 regression model. Best results among the proposed strategies are \textbf{bolded}, and second-best are \underline{underlined} (excluding the Original S2 Oracle).}
\label{table:supp_metrics_4panel}
\begin{tabular}{l c c c c}
\toprule
\textbf{Input Modality} & \textbf{Accuracy} & \textbf{Precision} & \textbf{Recall} & \textbf{Specificity} \\
\midrule
\multicolumn{5}{c}{\textbf{Panel A: Event Stratified Splitting (Cross-Validation Mean $\pm$ SD)}} \\
\midrule
S1 Only & 0.9699 $\pm$ 0.0103 & 0.8011 $\pm$ 0.0857 & 0.7456 $\pm$ 0.1856 & 0.9857 $\pm$ 0.0064 \\
S1 PCA & 0.9695 $\pm$ 0.0080 & 0.7835 $\pm$ 0.0887 & \underline{0.7910 $\pm$ 0.1049} & 0.9825 $\pm$ 0.0073 \\
S1 PCA + Raw S1 & \underline{0.9716 $\pm$ 0.0104} & \textbf{0.8229 $\pm$ 0.0745} & 0.7361 $\pm$ 0.1835 & \textbf{0.9878 $\pm$ 0.0045} \\
Synthetic S2 (Reg) & 0.9717 $\pm$ 0.0094 & \underline{0.8161 $\pm$ 0.0912} & 0.7576 $\pm$ 0.1108 & \underline{0.9868 $\pm$ 0.0058} \\
Combo (Synth S2 + S1) & \textbf{0.9741 $\pm$ 0.0070} & 0.8153 $\pm$ 0.0764 & \textbf{0.7972 $\pm$ 0.1108} & 0.9859 $\pm$ 0.0048 \\
\midrule
\textit{Original S2 (Oracle)} & \textit{0.9842 $\pm$ 0.0090} & \textit{0.9126 $\pm$ 0.0668} & \textit{0.8884 $\pm$ 0.0284} & \textit{0.9929 $\pm$ 0.0051} \\
\midrule
\multicolumn{5}{c}{\textbf{Panel B: Event Stratified Splitting (Global Micro-Aggregated)}} \\
\midrule
S1 Only & 0.9674 & 0.8623 & 0.8038 & 0.9857 \\
S1 PCA & 0.9682 & 0.8432 & \underline{0.8397} & 0.9825 \\
S1 PCA + Raw S1 & \underline{0.9692} & \textbf{0.8815} & 0.8020 & \textbf{0.9879} \\
Synthetic S2 (Reg) & 0.9697 & \underline{0.8693} & 0.8221 & \underline{0.9862} \\
Combo (Synth S2 + S1) & \textbf{0.9724} & 0.8670 & \textbf{0.8574} & 0.9853 \\
\midrule
\textit{Original S2 (Oracle)} & \textit{0.9828} & \textit{0.9274} & \textit{0.8998} & \textit{0.9921} \\
\midrule
\multicolumn{5}{c}{\textbf{Panel C: Mixed Splitting (Cross-Validation Mean $\pm$ SD)}} \\
\midrule
S1 Only & 0.9789 $\pm$ 0.0100 & 0.9175 $\pm$ 0.0085 & 0.8677 $\pm$ 0.0088 & 0.9913 $\pm$ 0.0007 \\
S1 PCA & 0.9784 $\pm$ 0.0018 & 0.9173 $\pm$ 0.0117 & 0.8622 $\pm$ 0.0203 & 0.9914 $\pm$ 0.0009 \\
S1 PCA + Raw S1 & 0.9789 $\pm$ 0.0012 & 0.9104 $\pm$ 0.0143 & 0.8762 $\pm$ 0.0039 & 0.9904 $\pm$ 0.0012 \\
Synthetic S2 (Reg) & \underline{0.9817 $\pm$ 0.0011} & \textbf{0.9325 $\pm$ 0.0078} & \underline{0.8805 $\pm$ 0.0106} & \textbf{0.9929 $\pm$ 0.0005} \\
Combo (Synth S2 + S1) & \textbf{0.9823 $\pm$ 0.0014} & \underline{0.9304 $\pm$ 0.0076} & \textbf{0.8896 $\pm$ 0.0143} & \underline{0.9925 $\pm$ 0.0010} \\
\midrule
\textit{Original S2 (Oracle)} & \textit{0.9862 $\pm$ 0.0005} & \textit{0.9522 $\pm$ 0.0054} & \textit{0.9078 $\pm$ 0.0069} & \textit{0.9949 $\pm$ 0.0003} \\
\midrule
\multicolumn{5}{c}{\textbf{Panel D: Mixed Splitting (Global Micro-Aggregated)}} \\
\midrule
S1 Only & 0.9789 & 0.9178 & 0.8678 & 0.9913 \\
S1 PCA & 0.9784 & 0.9178 & 0.8622 & 0.9914 \\
S1 PCA + Raw S1 & 0.9789 & 0.9109 & 0.8763 & 0.9904 \\
Synthetic S2 (Reg) & \underline{0.9817} & \textbf{0.9329} & \underline{0.8810} & \textbf{0.9929} \\
Combo (Synth S2 + S1) & \textbf{0.9823} & \underline{0.9304} & \textbf{0.8903} & \underline{0.9926} \\
\midrule
\textit{Original S2 (Oracle)} & \textit{0.9862} & \textit{0.9525} & \textit{0.9082} & \textit{0.9949} \\
\bottomrule
\end{tabular}
\end{table*}

\begin{table*}[!ht]
\centering
\caption{Comprehensive ablation study on architectural capacity and input tile size under both Event-Stratified and Mixed cross-validation frameworks. Performance is evaluated using Specificity and Recall across Cross-Validation (CV) and Global Micro-aggregated metrics. The configuration is denoted as \textbf{Reg-[X] / Seg-[Y]}, indicating the capacity of the regression and segmentation backbones, respectively. Optimal metrics are \textbf{bolded}; second-best values are \underline{underlined}.}
\label{tab:ablation_architecture_tiles_spec_recall}
\renewcommand{\arraystretch}{1.0} 
\setlength{\tabcolsep}{6pt}
\small 
\begin{tabular}{lll cccc}
\toprule
\multirow{2}{*}{\textbf{Tile Size}} & \multirow{2}{*}{\textbf{Configuration}} & \multirow{2}{*}{\textbf{Input Modality}} & \multicolumn{2}{c}{\textbf{Specificity}} & \multicolumn{2}{c}{\textbf{Recall}} \\
\cmidrule(lr){4-5} \cmidrule(lr){6-7}
& & & \textbf{CV (Mean $\pm$ Std)} & \textbf{Global} & \textbf{CV (Mean $\pm$ Std)} & \textbf{Global} \\
\midrule
\multicolumn{7}{c}{\textbf{Panel A: Event-Stratified Cross-Validation}} \\
\midrule
\multirow{12}{*}{$128 \times 128$} 
& \multirow{3}{*}{Reg-B0 / Seg-B0} 
  & S1 Only & \underline{0.9859 $\pm$ 0.0043} & \textbf{0.9854} & \underline{0.7698 $\pm$ 0.1551} & \underline{0.8234} \\
& & Syn S2 Only & $0.9853 \pm 0.0042$ & $0.9850$ & $0.6710 \pm 0.2817$ & $0.8162$ \\
& & Combined & \textbf{0.9860 $\pm$ 0.0062} & \underline{0.9851} & \textbf{0.7904 $\pm$ 0.0964} & \textbf{0.8503} \\
\cmidrule{2-7}
& \multirow{3}{*}{Reg-B3 / Seg-B0} 
  & S1 Only & $0.9848 \pm 0.0071$ & \underline{0.9848} & \underline{0.7218 $\pm$ 0.1696} & \underline{0.7971} \\
& & Syn S2 Only & \textbf{0.9881 $\pm$ 0.0057} & \textbf{0.9882} & $0.7003 \pm 0.1441$ & $0.7944$ \\
& & Combined & \underline{0.9849 $\pm$ 0.0085} & $0.9846$ & \textbf{0.7669 $\pm$ 0.1286} & \textbf{0.8292} \\
\cmidrule{2-7}
& \multirow{3}{*}{Reg-B3 / Seg-B3} 
  & S1 Only & $0.9811 \pm 0.0087$ & $0.9790$ & \underline{0.7539 $\pm$ 0.1347} & \textbf{0.8095} \\
& & Syn S2 Only & \textbf{0.9885 $\pm$ 0.0054} & \textbf{0.9881} & $0.6895 \pm 0.1554$ & $0.7768$ \\
& & Combined & \underline{0.9843 $\pm$ 0.0075} & \underline{0.9847} & \textbf{0.7644 $\pm$ 0.1646} & \underline{0.8073} \\
\cmidrule{2-7}
& \multirow{3}{*}{Reg-B5 / Seg-B0} 
  & S1 Only & $0.9857 \pm 0.0064$ & \underline{0.9857} & $0.7456 \pm 0.1856$ & $0.8038$ \\
& & Syn S2 Only & \textbf{0.9868 $\pm$ 0.0058} & \textbf{0.9862} & \underline{0.7576 $\pm$ 0.1108} & \underline{0.8221} \\
& & Combined & \underline{0.9859 $\pm$ 0.0048} & $0.9853$ & \textbf{0.7972 $\pm$ 0.1108} & \textbf{0.8574} \\

\midrule
\multirow{9}{*}{$256 \times 256$} 
& \multirow{3}{*}{Reg-B0 / Seg-B0} 
  & S1 Only & $0.9744 \pm 0.0170$ & $0.9691$ & \textbf{0.6232 $\pm$ 0.1644} & \textbf{0.6858} \\
& & Syn S2 Only & \textbf{0.9903 $\pm$ 0.0028} & \textbf{0.9914} & $0.4922 \pm 0.2536$ & $0.5871$ \\
& & Combined & \underline{0.9791 $\pm$ 0.0181} & \underline{0.9728} & \underline{0.5925 $\pm$ 0.2377} & \underline{0.6737} \\
\cmidrule{2-7}
& \multirow{3}{*}{Reg-B3 / Seg-B0} 
  & S1 Only & $0.9470 \pm 0.0542$ & $0.9637$ & $0.6397 \pm 0.2134$ & $0.7357$ \\
& & Syn S2 Only & \underline{0.9751 $\pm$ 0.0202} & \underline{0.9804} & \textbf{0.6695 $\pm$ 0.2312} & \textbf{0.7779} \\
& & Combined & \textbf{0.9802 $\pm$ 0.0083} & \textbf{0.9816} & \underline{0.6662 $\pm$ 0.2027} & \underline{0.7677} \\
\cmidrule{2-7}
& \multirow{3}{*}{Reg-B5 / Seg-B0} 
  & S1 Only & $0.9673 \pm 0.0220$ & $0.9696$ & \underline{0.6534 $\pm$ 0.2247} & \underline{0.7416} \\
& & Syn S2 Only & \textbf{0.9798 $\pm$ 0.0081} & \textbf{0.9811} & $0.6202 \pm 0.2391$ & $0.7379$ \\
& & Combined & \underline{0.9770 $\pm$ 0.0091} & \underline{0.9800} & \textbf{0.6718 $\pm$ 0.1895} & \textbf{0.7708} \\

\midrule
\multicolumn{7}{c}{\textbf{Panel B: Mixed Cross-Validation}} \\
\midrule
\multirow{12}{*}{$128 \times 128$} 
& \multirow{3}{*}{Reg-B0 / Seg-B0} 
  & S1 Only & $0.9919 \pm 0.0012$ & $0.9919$ & $0.8653 \pm 0.0183$ & $0.8659$ \\
& & Syn S2 Only & \textbf{0.9930 $\pm$ 0.0008} & \textbf{0.9930} & \underline{0.8672 $\pm$ 0.0103} & \underline{0.8678} \\
& & Combined & \underline{0.9924 $\pm$ 0.0012} & \underline{0.9924} & \textbf{0.8793 $\pm$ 0.0127} & \textbf{0.8800} \\
\cmidrule{2-7}
& \multirow{3}{*}{Reg-B3 / Seg-B0} 
  & S1 Only & $0.9906 \pm 0.0006$ & $0.9906$ & $0.8659 \pm 0.0269$ & $0.8672$ \\
& & Syn S2 Only & \textbf{0.9929 $\pm$ 0.0006} & \textbf{0.9929} & \underline{0.8762 $\pm$ 0.0122} & \underline{0.8767} \\
& & Combined & \underline{0.9925 $\pm$ 0.0009} & \underline{0.9925} & \textbf{0.8868 $\pm$ 0.0184} & \textbf{0.8875} \\
\cmidrule{2-7}
& \multirow{3}{*}{Reg-B3 / Seg-B3} 
  & S1 Only & $0.9924 \pm 0.0005$ & $0.9924$ & $0.8708 \pm 0.0211$ & $0.8718$ \\
& & Syn S2 Only & \underline{0.9925 $\pm$ 0.0007} & \underline{0.9925} & \underline{0.8805 $\pm$ 0.0116} & \underline{0.8809} \\
& & Combined & \textbf{0.9928 $\pm$ 0.0009} & \textbf{0.9928} & \textbf{0.8899 $\pm$ 0.0084} & \textbf{0.8901} \\
\cmidrule{2-7}
& \multirow{3}{*}{Reg-B5 / Seg-B0} 
  & S1 Only & $0.9913 \pm 0.0007$ & $0.9913$ & $0.8677 \pm 0.0088$ & $0.8678$ \\
& & Syn S2 Only & \textbf{0.9929 $\pm$ 0.0005} & \textbf{0.9929} & \underline{0.8805 $\pm$ 0.0106} & \underline{0.8810} \\
& & Combined & \underline{0.9925 $\pm$ 0.0010} & \underline{0.9926} & \textbf{0.8896 $\pm$ 0.0143} & \textbf{0.8903} \\

\midrule
\multirow{15}{*}{$256 \times 256$} 
& \multirow{3}{*}{Reg-B0 / Seg-B0} 
  & S1 Only & $0.9862 \pm 0.0027$ & $0.9862$ & \underline{0.7947 $\pm$ 0.0155} & \underline{0.7959} \\
& & Syn S2 Only & \textbf{0.9916 $\pm$ 0.0011} & \textbf{0.9916} & $0.7838 \pm 0.0342$ & $0.7877$ \\
& & Combined & \underline{0.9907 $\pm$ 0.0022} & \underline{0.9907} & \textbf{0.7974 $\pm$ 0.0392} & \textbf{0.8024} \\
\cmidrule{2-7}
& \multirow{3}{*}{Reg-B2 / Seg-B0} 
  & S1 Only & $0.9872 \pm 0.0000$ & $0.9872$ & \textbf{0.8239 $\pm$ 0.0000} & \textbf{0.8239} \\
& & Syn S2 Only & \underline{0.9905 $\pm$ 0.0000} & \underline{0.9905} & $0.7858 \pm 0.0000$ & $0.7858$ \\
& & Combined & \textbf{0.9905 $\pm$ 0.0000} & \textbf{0.9905} & \underline{0.8051 $\pm$ 0.0000} & \underline{0.8051} \\
\cmidrule{2-7}
& \multirow{3}{*}{Reg-B3 / Seg-B0} 
  & S1 Only & $0.9871 \pm 0.0018$ & $0.9871$ & \underline{0.8135 $\pm$ 0.0294} & \underline{0.8156} \\
& & Syn S2 Only & \textbf{0.9917 $\pm$ 0.0025} & \textbf{0.9918} & $0.8057 \pm 0.0598$ & $0.8135$ \\
& & Combined & \underline{0.9909 $\pm$ 0.0023} & \underline{0.9910} & \textbf{0.8248 $\pm$ 0.0551} & \textbf{0.8323} \\
\cmidrule{2-7}
& \multirow{3}{*}{Reg-B3 / Seg-B3} 
  & S1 Only & $0.9849 \pm 0.0031$ & $0.9849$ & \underline{0.8306 $\pm$ 0.0181} & \underline{0.8328} \\
& & Syn S2 Only & \textbf{0.9882 $\pm$ 0.0029} & \textbf{0.9882} & $0.8148 \pm 0.0447$ & $0.8201$ \\
& & Combined & \underline{0.9882 $\pm$ 0.0023} & \underline{0.9882} & \textbf{0.8374 $\pm$ 0.0352} & \textbf{0.8415} \\
\cmidrule{2-7}
& \multirow{3}{*}{Reg-B5 / Seg-B0} 
  & S1 Only & $0.9863 \pm 0.0044$ & $0.9863$ & \underline{0.8112 $\pm$ 0.0240} & \underline{0.8141} \\
& & Syn S2 Only & \underline{0.9913 $\pm$ 0.0015} & \underline{0.9913} & $0.8013 \pm 0.0453$ & $0.8056$ \\
& & Combined & \textbf{0.9914 $\pm$ 0.0022} & \textbf{0.9914} & \textbf{0.8179 $\pm$ 0.0321} & \textbf{0.8208} \\

\bottomrule
\end{tabular}
\end{table*}

\begin{table*}[!ht]
\centering
\caption{Comprehensive ablation study on architectural capacity and input tile size under both Event-Stratified and Mixed cross-validation frameworks. Performance is evaluated using Precision and Accuracy across Cross-Validation (CV) and Global Micro-aggregated metrics. The configuration is denoted as \textbf{Reg-[X] / Seg-[Y]}, indicating the capacity of the regression and segmentation backbones, respectively. Optimal metrics are \textbf{bolded}; second-best values are \underline{underlined}.}
\label{tab:ablation_architecture_tiles_prec_acc}
\renewcommand{\arraystretch}{1.0} 
\setlength{\tabcolsep}{6pt}
\small 
\begin{tabular}{lll cccc}
\toprule
\multirow{2}{*}{\textbf{Tile Size}} & \multirow{2}{*}{\textbf{Configuration}} & \multirow{2}{*}{\textbf{Input Modality}} & \multicolumn{2}{c}{\textbf{Precision}} & \multicolumn{2}{c}{\textbf{Accuracy}} \\
\cmidrule(lr){4-5} \cmidrule(lr){6-7}
& & & \textbf{CV (Mean $\pm$ Std)} & \textbf{Global} & \textbf{CV (Mean $\pm$ Std)} & \textbf{Global} \\
\midrule
\multicolumn{7}{c}{\textbf{Panel A: Event-Stratified Cross-Validation}} \\
\midrule
\multirow{12}{*}{$128 \times 128$} 
& \multirow{3}{*}{Reg-B0 / Seg-B0} 
  & S1 Only & \underline{0.7957 $\pm$ 0.0913} & \underline{0.8633} & \underline{0.9716 $\pm$ 0.0101} & \underline{0.9691} \\
& & Syn S2 Only & $0.7111 \pm 0.2346$ & $0.8589$ & $0.9683 \pm 0.0055$ & $0.9680$ \\
& & Combined & \textbf{0.8119 $\pm$ 0.0895} & \textbf{0.8645} & \textbf{0.9733 $\pm$ 0.0074} & \textbf{0.9715} \\
\cmidrule{2-7}
& \multirow{3}{*}{Reg-B3 / Seg-B0} 
  & S1 Only & $0.7963 \pm 0.0808$ & $0.8544$ & $0.9681 \pm 0.0098$ & $0.9659$ \\
& & Syn S2 Only & \textbf{0.8302 $\pm$ 0.0615} & \textbf{0.8830} & \underline{0.9701 $\pm$ 0.0072} & \underline{0.9687} \\
& & Combined & \underline{0.8138 $\pm$ 0.0847} & \underline{0.8579} & \textbf{0.9708 $\pm$ 0.0088} & \textbf{0.9690} \\
\cmidrule{2-7}
& \multirow{3}{*}{Reg-B3 / Seg-B3} 
  & S1 Only & $0.7483 \pm 0.1366$ & $0.8114$ & $0.9650 \pm 0.0138$ & $0.9619$ \\
& & Syn S2 Only & \textbf{0.8184 $\pm$ 0.0834} & \textbf{0.8791} & \underline{0.9686 $\pm$ 0.0091} & \underline{0.9668} \\
& & Combined & \underline{0.7835 $\pm$ 0.0983} & \underline{0.8547} & \textbf{0.9688 $\pm$ 0.0087} & \textbf{0.9668} \\
\cmidrule{2-7}
& \multirow{3}{*}{Reg-B5 / Seg-B0} 
  & S1 Only & $0.8011 \pm 0.0857$ & $0.8623$ & $0.9699 \pm 0.0103$ & $0.9674$ \\
& & Syn S2 Only & \textbf{0.8161 $\pm$ 0.0912} & \textbf{0.8693} & \underline{0.9717 $\pm$ 0.0094} & \underline{0.9697} \\
& & Combined & \underline{0.8153 $\pm$ 0.0764} & \underline{0.8670} & \textbf{0.9741 $\pm$ 0.0070} & \textbf{0.9724} \\

\midrule
\multirow{9}{*}{$256 \times 256$} 
& \multirow{3}{*}{Reg-B0 / Seg-B0} 
  & S1 Only & $0.6378 \pm 0.1569$ & $0.5439$ & $0.9484 \pm 0.0177$ & $0.9546$ \\
& & Syn S2 Only & \textbf{0.7251 $\pm$ 0.1140} & \textbf{0.7862} & \textbf{0.9534 $\pm$ 0.0310} & \textbf{0.9708} \\
& & Combined & \underline{0.6871 $\pm$ 0.1650} & \underline{0.5712} & \underline{0.9495 $\pm$ 0.0185} & \underline{0.9575} \\
\cmidrule{2-7}
& \multirow{3}{*}{Reg-B3 / Seg-B0} 
  & S1 Only & $0.6844 \pm 0.1734$ & $0.6920$ & $0.9343 \pm 0.0198$ & $0.9410$ \\
& & Syn S2 Only & \textbf{0.7758 $\pm$ 0.1177} & \underline{0.8147} & \underline{0.9545 $\pm$ 0.0119} & \underline{0.9602} \\
& & Combined & \underline{0.7719 $\pm$ 0.1080} & \textbf{0.8220} & \textbf{0.9555 $\pm$ 0.0137} & \textbf{0.9603} \\
\cmidrule{2-7}
& \multirow{3}{*}{Reg-B5 / Seg-B0} 
  & S1 Only & $0.7088 \pm 0.1618$ & $0.7298$ & $0.9444 \pm 0.0131$ & $0.9469$ \\
& & Syn S2 Only & \underline{0.7438 $\pm$ 0.1290} & \textbf{0.8119} & \underline{0.9482 $\pm$ 0.0185} & \underline{0.9568} \\
& & Combined & \textbf{0.7493 $\pm$ 0.1285} & \underline{0.8103} & \textbf{0.9520 $\pm$ 0.0188} & \textbf{0.9591} \\

\midrule
\multicolumn{7}{c}{\textbf{Panel B: Mixed Cross-Validation}} \\
\midrule
\multirow{12}{*}{$128 \times 128$} 
& \multirow{3}{*}{Reg-B0 / Seg-B0} 
  & S1 Only & $0.9227 \pm 0.0097$ & $0.9227$ & $0.9792 \pm 0.0016$ & $0.9792$ \\
& & Syn S2 Only & \textbf{0.9327 $\pm$ 0.0087} & \textbf{0.9329} & \underline{0.9804 $\pm$ 0.0009} & \underline{0.9804} \\
& & Combined & \underline{0.9287 $\pm$ 0.0105} & \underline{0.9287} & \textbf{0.9811 $\pm$ 0.0014} & \textbf{0.9811} \\
\cmidrule{2-7}
& \multirow{3}{*}{Reg-B3 / Seg-B0} 
  & S1 Only & $0.9112 \pm 0.0099$ & $0.9118$ & $0.9782 \pm 0.0022$ & $0.9782$ \\
& & Syn S2 Only & \textbf{0.9320 $\pm$ 0.0064} & \textbf{0.9322} & \underline{0.9812 $\pm$ 0.0011} & \underline{0.9812} \\
& & Combined & \underline{0.9299 $\pm$ 0.0053} & \underline{0.9299} & \textbf{0.9820 $\pm$ 0.0015} & \textbf{0.9820} \\
\cmidrule{2-7}
& \multirow{3}{*}{Reg-B3 / Seg-B3} 
  & S1 Only & $0.9285 \pm 0.0083$ & $0.9289$ & $0.9801 \pm 0.0019$ & $0.9801$ \\
& & Syn S2 Only & \underline{0.9302 $\pm$ 0.0074} & \underline{0.9304} & \underline{0.9811 $\pm$ 0.0007} & \underline{0.9811} \\
& & Combined & \textbf{0.9335 $\pm$ 0.0117} & \textbf{0.9340} & \textbf{0.9823 $\pm$ 0.0010} & \textbf{0.9823} \\
\cmidrule{2-7}
& \multirow{3}{*}{Reg-B5 / Seg-B0} 
  & S1 Only & $0.9175 \pm 0.0085$ & $0.9178$ & $0.9789 \pm 0.0010$ & $0.9789$ \\
& & Syn S2 Only & \textbf{0.9325 $\pm$ 0.0078} & \textbf{0.9329} & \underline{0.9817 $\pm$ 0.0011} & \underline{0.9817} \\
& & Combined & \underline{0.9304 $\pm$ 0.0076} & \underline{0.9304} & \textbf{0.9823 $\pm$ 0.0014} & \textbf{0.9823} \\

\midrule
\multirow{15}{*}{$256 \times 256$} 
& \multirow{3}{*}{Reg-B0 / Seg-B0} 
  & S1 Only & $0.8658 \pm 0.0266$ & $0.8675$ & $0.9667 \pm 0.0036$ & $0.9667$ \\
& & Syn S2 Only & \textbf{0.9132 $\pm$ 0.0086} & \textbf{0.9144} & \underline{0.9708 $\pm$ 0.0021} & \underline{0.9708} \\
& & Combined & \underline{0.9079 $\pm$ 0.0109} & \underline{0.9074} & \textbf{0.9715 $\pm$ 0.0012} & \textbf{0.9715} \\
\cmidrule{2-7}
& \multirow{3}{*}{Reg-B2 / Seg-B0} 
  & S1 Only & $0.8775 \pm 0.0000$ & $0.8775$ & \underline{0.9708 $\pm$ 0.0000} & \underline{0.9708} \\
& & Syn S2 Only & \underline{0.9022 $\pm$ 0.0000} & \underline{0.9022} & $0.9699 \pm 0.0000$ & $0.9699$ \\
& & Combined & \textbf{0.9047 $\pm$ 0.0000} & \textbf{0.9047} & \textbf{0.9719 $\pm$ 0.0000} & \textbf{0.9719} \\
\cmidrule{2-7}
& \multirow{3}{*}{Reg-B3 / Seg-B0} 
  & S1 Only & $0.8725 \pm 0.0212$ & $0.8746$ & $0.9700 \pm 0.0045$ & $0.9700$ \\
& & Syn S2 Only & \textbf{0.9165 $\pm$ 0.0161} & \textbf{0.9160} & \underline{0.9740 $\pm$ 0.0038} & \underline{0.9740} \\
& & Combined & \underline{0.9100 $\pm$ 0.0152} & \underline{0.9106} & \textbf{0.9752 $\pm$ 0.0030} & \textbf{0.9752} \\
\cmidrule{2-7}
& \multirow{3}{*}{Reg-B3 / Seg-B3} 
  & S1 Only & $0.8583 \pm 0.0264$ & $0.8594$ & $0.9697 \pm 0.0031$ & $0.9697$ \\
& & Syn S2 Only & \underline{0.8826 $\pm$ 0.0312} & \underline{0.8850} & \underline{0.9714 $\pm$ 0.0038} & \underline{0.9714} \\
& & Combined & \textbf{0.8855 $\pm$ 0.0240} & \textbf{0.8874} & \textbf{0.9735 $\pm$ 0.0029} & \textbf{0.9735} \\
\cmidrule{2-7}
& \multirow{3}{*}{Reg-B5 / Seg-B0} 
  & S1 Only & $0.8695 \pm 0.0287$ & $0.8684$ & $0.9692 \pm 0.0039$ & $0.9692$ \\
& & Syn S2 Only & \underline{0.9101 $\pm$ 0.0142} & \underline{0.9113} & \underline{0.9728 $\pm$ 0.0038} & \underline{0.9728} \\
& & Combined & \textbf{0.9124 $\pm$ 0.0210} & \textbf{0.9134} & \textbf{0.9744 $\pm$ 0.0031} & \textbf{0.9744} \\

\bottomrule
\end{tabular}
\end{table*}

\end{document}